\documentclass[preprint,12pt]{elsarticle}
\usepackage{subcaption}
\usepackage{graphicx}
\newcommand{\bb}{\mathbf}
\usepackage{url}
\usepackage{float}
\usepackage{amsmath}
\usepackage[ruled,vlined]{algorithm2e}

\usepackage{amssymb}

\begin{document}

\begin{frontmatter}

\title{AquaFeL-PSO: A Monitoring System for Water Resources using Autonomous Surface Vehicles based on Multimodal PSO and Federated Learning}

\author[inst1]{Micaela Jara Ten Kathen}

\affiliation[inst1]{organization={Department of Engineering, Universidad Loyola},
            addressline={Av. de las Universidades, s/n}, 
            city={Dos Hermanas},
            postcode={41704}, 
            state={Seville},
            country={Spain}}

\author[inst2]{Princy Johnson}

\affiliation[inst2]{organization={School of Engineering, Liverpool John Moores University},
            addressline={Byrom St.}, 
            city={Liverpool},
            postcode={L3 3AF}, 
            state={Merseyside},
            country={United Kingdom}}

\author[inst1]{Isabel Jurado Flores}

\author[inst3]{Daniel Guti{\'e}rrez Reina\corref{cor1}}
\ead{dgutierrezreina@us.es}
\cortext[cor1]{Corresponding author}

\affiliation[inst3]{organization={Department of Electronic Engineering, Technical School of Engineering of Seville},
            addressline={C. Americo Vespucio}, 
            city={Seville},
            postcode={41092}, 
            state={Seville},
            country={Spain}}

\begin{abstract}
    The preservation, monitoring, and control of water resources has been a major challenge in recent decades. Water resources must be constantly monitored to know the contamination levels of water. To meet this objective, this paper proposes a water monitoring system using autonomous surface vehicles, equipped with water quality sensors, based on a multimodal particle swarm optimization, and the federated learning technique, with Gaussian process as a surrogate model, the AquaFeL-PSO algorithm. The proposed monitoring system has two phases, the exploration phase and the exploitation phase. In the exploration phase, the vehicles examine the surface of the water resource, and with the data acquired by the water quality sensors, a first water quality model is estimated in the central server. In the exploitation phase, the area is divided into action zones using the model estimated in the exploration phase for a better exploitation of the contamination zones. To obtain the final water quality model of the water resource, the models obtained in both phases are combined. The results demonstrate the efficiency of the proposed path planner in obtaining water quality models of the pollution zones, with a 14$\%$ improvement over the other path planners compared, and the entire water resource, obtaining a 400$\%$ better model, as well as in detecting pollution peaks, the improvement in this case study is 4,000$\%$. It was also proven that the results obtained by applying the federated learning technique are very similar to the results of a centralized system.
\end{abstract}

\begin{graphicalabstract}
    \begin{figure}[ht!]
        \centering
        \includegraphics[width=\textwidth]{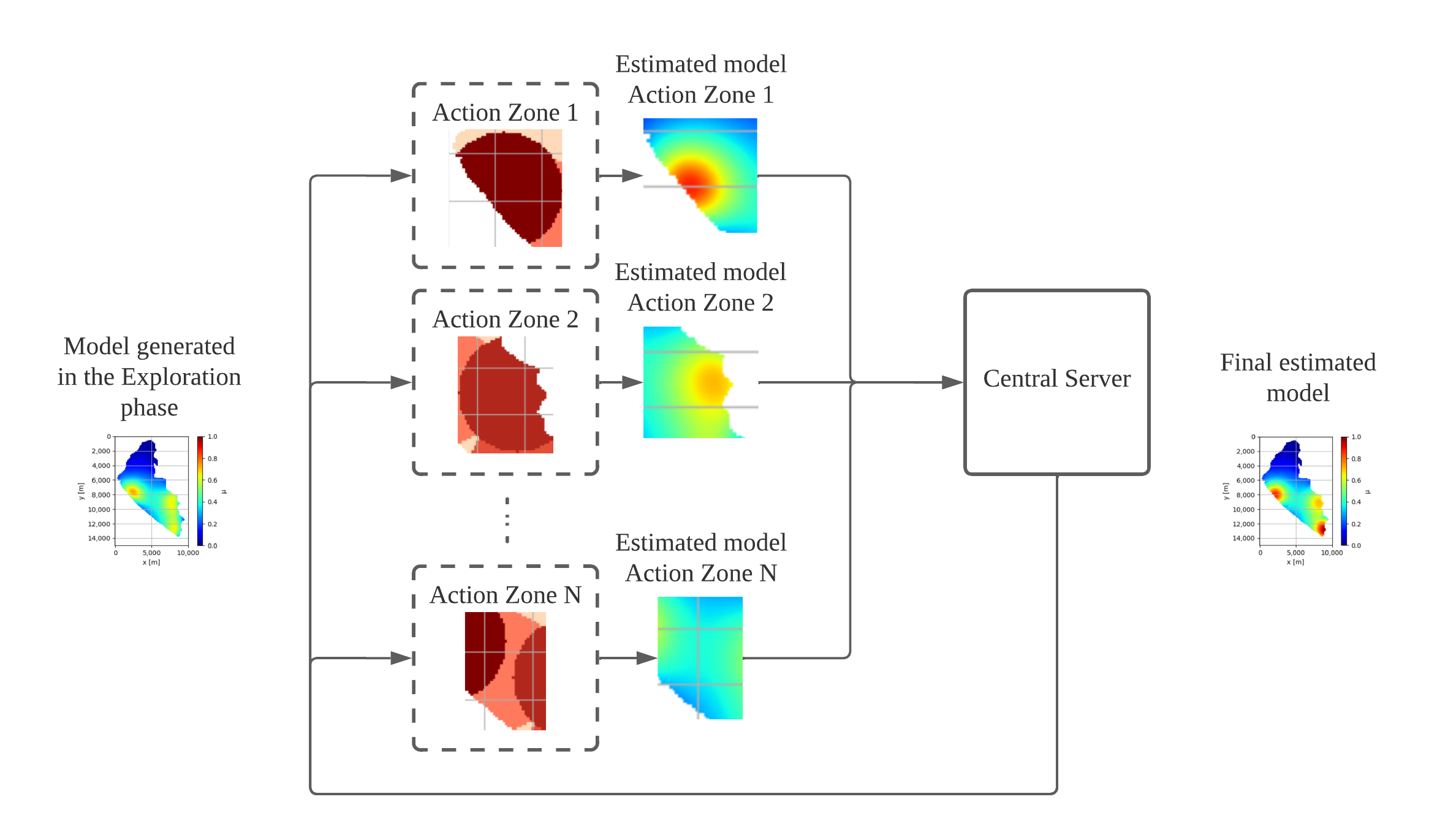}
    \end{figure}
    \vfill
\end{graphicalabstract}

\begin{highlights}
    \item \raggedright An informative path planning algorithm based on Particle Swarm Optimization, Gaussian Process, and Federated Learning
    \item \raggedright AquaFeL-PSO monitoring system achieves satisfactory results in multimodal water resource monitoring problems
    \item \raggedright AquaFeL-PSO outperforms significantly other counterparts such as classical PSO, PSO-based on surrogate models, and lawnmower algorithm, in terms of mean square error.   
    \item \raggedright Federated learning approach has been validated for water estimation models through autonomous vehicles.
\end{highlights}


\end{frontmatter}

\section{Introduction}
\label{sec:introduction}

Water is the source of life. Although it is a renewable resource, it is becoming increasingly scarce worldwide. Fresh water reservoirs are being polluted by domestic and industrial effluents and fertilizers from agricultural activities, among other causes. These problems not only affect the quality of life of human beings but also affect the planet. In 2015, at the United Nations Summit, the 2030 Agenda for Sustainable Development was approved\footnote{https://sdgs.un.org/es/goals}. The said document sets out 17 sustainable development goals and 169 targets to help people, the planet and prosperity. Among the 17 goals, the sixth one has water and sanitation as its main objectives. Reducing pollution, efficient use of water resources, water management, and the development of water-related programs are some of the pathways to achieve this goal \cite{desa2016transforming}. 

A necessary method to control the progress of actions taken to reduce pollution in water resources is monitoring the water quality of the water body. Water bodies that are constantly threatened must be monitored on a permanent basis. A prime example where contamination is always being studied and monitored is the Ypacarai lake in Paraguay \cite{itaipu2018}\cite{itaipu2021}. The main cause of contamination is the domestic and industrial effluents that flow into the lake and its watersheds \cite{lopez2018eutrophication}. As a consequence, the levels of fecal coliforms and nutrients such as phosphorus and nitrate are high. High levels of nutrients cause the appearance of cyanobacteria \cite{arzamendia2019intelligent}. This problem affects public health, the economy of the region, and the ecosystem in and around the lake.

\begin{figure}[H]
    \centering
    \includegraphics[width=0.5\textwidth]{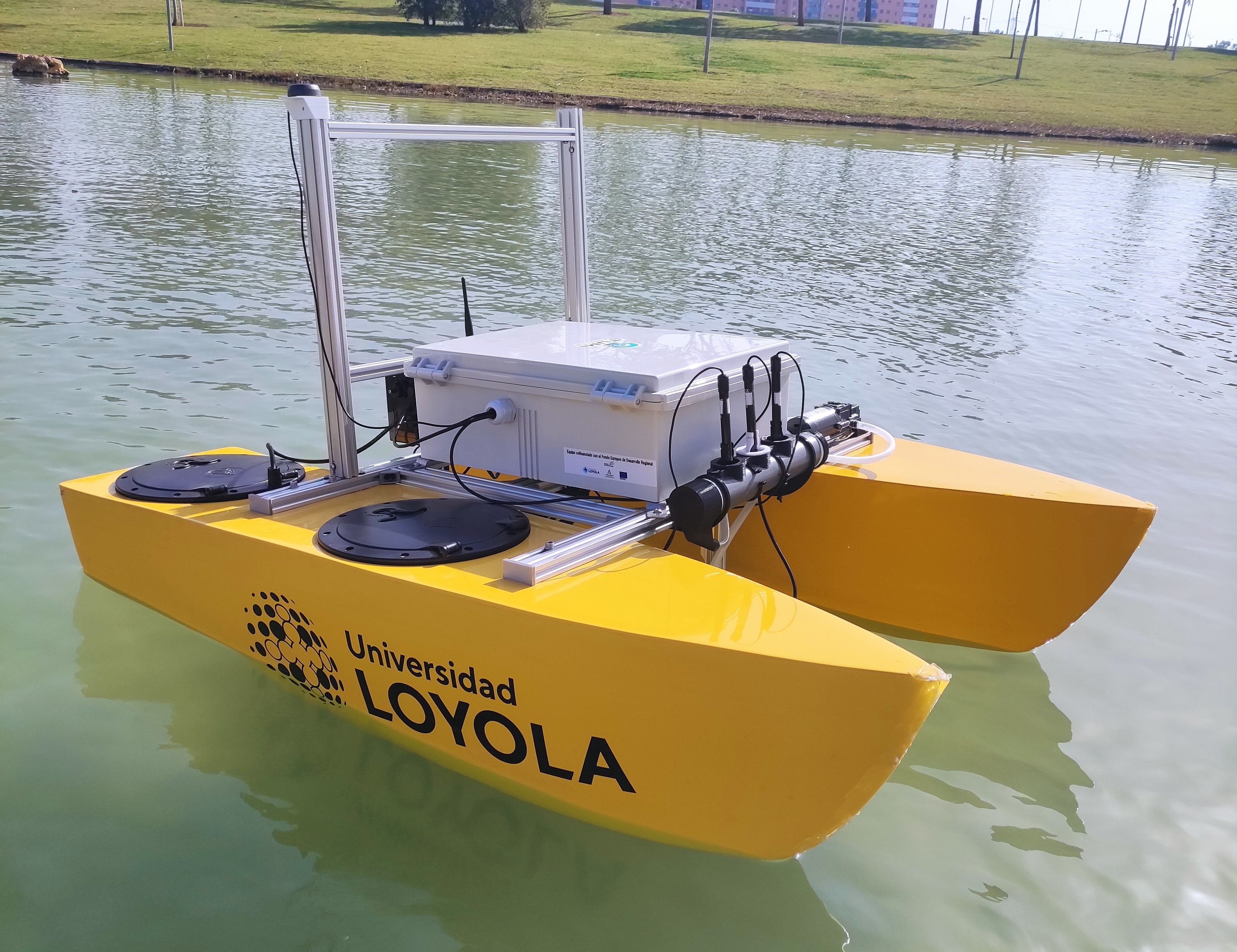}
    \caption{An Autonomous Surface Vehicle used for water resources monitoring system}
    \label{fig:asv}
\end{figure}

The traditional method of monitoring water quality consists of taking water samples at specific points in the water resource and analyzing those samples in specialized laboratories. This method can be exhaustive, time-consuming, human error can occur, and the costs of hiring personnel and equipment can be high \cite{luis2021multiagent}. New technologies, such as aquatic submarines, surface vehicles, or aerial drones equipped with sensors capable of measuring water parameters, are currently being widely studied \cite{lin2022smart}\cite{peralta2021}\cite{luis2021multiagent}\cite{panetsos2022vision}. Vehicles are equipped with Guidance, Navigation, and Control (GNC) systems that allow vehicles to move autonomously throughout the water body. One of the vehicles of the Autonomous Surface Vehicles (ASVs) fleet is shown in Fig.~\ref{fig:asv}. Control and navigation tests have been conducted at Lago de la Vida (Dos Hermanas, Spain)\footnote{\url{https://www.youtube.com/watch?v=3nYzSSGeyHw}} and the Alamillo Park lake (Seville, Spain).

This work focuses on the guidance system. The guidance system is responsible for generating the path that autonomous vehicles have to travel, and the speed at which they have to go. One such algorithm that is being widely studied and used as a path planner is the Particle Swarm Optimization (PSO) \cite{sanchez2019distributed}\cite{song2017global}. The PSO is a Swarm Intelligence (SI) algorithm that is able to work with several agents/particles. The PSO is a simple metaheuristic algorithm to implement and does not have many parameters that need to be adjusted. The disadvantages of the PSO are the difficulty of initialing the configuration parameters and the ease with which particles get stuck in a local optimum. The monitoring system of a water resource can be considered a multimodal problem, because several pollution zones or several pollution peaks can be found in the water resource, whether global or local. 
Because of this, a solution is to develop an informative path planner. The informative path planner is a type of path planner that composes the guidance system. The informative path planner has the advantage that it makes use of the water samples taken during the travel of the ASVs for the generation of the path. With the data obtained, the monitoring system can generate an estimated water quality model using machine learning techniques \cite{bottarelli2019orienteering}. In this work, a multimodal PSO-based informative path planning is developed for ASVs, using Gaussian Process (GP) as the underlying model. The monitoring task is divided into two phases, the exploration phase and the exploitation phase. In the exploration phase, ASVs aim to cover the largest possible area of the water resource without focusing on a specific area, to obtain a reliable model of the water resource. In this phase, the centralized learning technique is applied. Obtaining this first model, the monitoring system moves on to the exploitation phase. In this second phase, action zones are determined, and these action zones are the areas of high levels of pollution. After determining the action zones, the vehicles are assigned to the zones detected in the exploration phase to exploit them in a distributed way and generate a second water resource model, the final estimated model. The model obtained in the exploitation phase is based on the data collected in the exploration phase plus the data acquired by the ASV(s) while exploiting the given action zone. As a consequence, a more detailed estimated water quality model is obtained. The main objective of this second phase is to obtain a good characterization of the contamination zones that were determined in the exploration phase. In the exploitation phase, the Federated Learning technique (FL) is used. In recent years, the FL technique has received a lot of attention and many studies have been conducted \cite{yang2019federated}\cite{chen2020wireless}\cite{ye2020federated}. FL is a special category of Distributed Machine Learning (DML) \cite{yang2019federated}. This technique alleviates the problem of heavy traffic of data in the central server by training the GP in the nodes of each action zone with data collected by the ASVs assigned to that area in the exploitation phase, by using the data from the exploration phase.

The main contributions of this work are:

\begin{itemize}
    \item The development of a monitoring system based on multimodal particle swarm optimization and Gaussian process for autonomous surface vehicles that is capable of generating accurate and precise estimated water quality models, and is able to detect areas of water resource contamination through autonomous vehicles.
    \item The comparison of centralized and federated learning for multimodal estimation of water quality parameters in water bodies.
    \item The application of the monitoring system for the case of Ypacarai lake, demonstrating in simulation the efficiency and effectiveness of the proposed approach over other path planners.
\end{itemize}

The paper layout is as follows: in Section~\ref{sec:state}, the state of the art of path planners for ASVs and PSO-based algorithms for multimodal problem solving are presented. The monitoring problem as well as the assumptions made to carry out the monitoring tasks are explained in Section~\ref{sec:statement}. Section~\ref{sec:psobased} contains the algorithms that are compared with the proposed approach. The proposed monitoring system is presented in Section~\ref{sec:proposed}. Section~\ref{sec:results} presents the simulation setup, evaluation of the proposed approach, and comparison of the performances of the path planners. Section~\ref{sec:disc} provides a discussion of the results. Conclusions of the work and future research directions are presented in Section~\ref{sec:conclusion}. 

\section{Related work}
\label{sec:state}

This state of the art section has been divided into two parts. First, similar approaches based on ASV and artificial intelligence algorithms for water monitoring are described. Followed by the review of the main techniques employed in PSO algorithm for multimodal optimization. 

\subsection{Water monitoring system based on ASVs}

In recent years, many studies of ASVs have been developed for application in areas that include the monitoring and patrolling of water resources \cite{arzamendia2019intelligent}\cite{peralta2021bayesian}\cite{luis2021multiagent}. To fulfill the task of monitoring water bodies, the vehicles are equipped with sensors capable of measuring water quality parameters such as ammonium, pH, temperature, among other parameters, as well as a Global Positioning System (GPS), an Inertial Measurement Unit (IMU), and/or a camera. For the development of the ASV guidance system, artificial intelligence techniques, such as Genetic Algorithm (GA) \cite{arzamendia2019intelligent}, Swarm Intelligence (SI) \cite{kathen2021informative}, Bayesian Optimization (BO) \cite{peralta2021}, Deep Reinforcement Learning (DRL) \cite{luis2020deep}, etc., are promising solutions.

The GA is a metaheuristic algorithm based on Charles Darwin's natural selection \cite{holland1992genetic}. This algorithm has been used as a path planner to solve the monitoring problem in \cite{arzamendia2019evolutionary}. This problem was modeled as the Traveling Salesman Problem (TSP), in which an ASV must cover the largest possible area of the water resource. For this purpose, the authors have fixed beacons on the shore of the water body. In \cite{arzamendia2019comparison}, the authors extended the work by modifying the modeling of the monitoring problem. In this extension, the problem was modeled as the Chinese Postman Problem (CPP). The difference between the TSP and the CPP is that, in the latter, the vehicle can pass more than once through the beacons. As a consequence, this allows maximizing the exploration of the lake surface. A two-phase GA-based monitoring system was developed in \cite{arzamendia2019intelligent}. In the first phase, the vehicle focuses on exploring the lake surface with the objective of detecting areas of contamination. In the second phase, the ASV focuses on exploiting the contamination zones found. 

DRL and BO are also used to solve water resources monitoring problems. In \cite{luis2020deep}, the monitoring problem was modeled as a Markov Decision Problem. The authors have developed a DRL-based online global path planner for ASVs to patrol the Ypacarai lake. RGB images of the water resource are used to represent possible vehicle positions and states. The authors trained a Deep Q-learning method based on convolutional neural networks using a tailored reward function. In \cite{luis2021multiagent}, the authors developed a multiagent monitoring system and used a centralized approach technique as an extension of their earlier work. The fleet was trained using a single neural network. The results showed that for the directed patrolling problem, the centralized approach is more efficient than distributed Q-learning. A comparison between GAs and DRL for monitoring problems through ASV can be found in \cite{yanes2021dimensional}. The authors demonstrated that in a complex scenario, DRL achieves better results. However, for low dimensional scenarios, GAs can obtain more robust solutions. In \cite{peralta2021}, the authors develop an informative path planning based on BO. The authors adapted the classical BO framework for water monitoring. The objective is to generate an estimated model of pollution in the Ypacarai lake. The authors compare different acquisition functions to determine the positions where water samples should be taken. In \cite{peralta2021bayesian}, the authors extended the work by considering the monitoring problem as a multiobjective problem where several water quality parameters were taken into account simultaneously.  In order to obtain the optimal positions for water sampling, the authors evaluated different multiobjective techniques to create an acquisition function that combines data from the water quality sensors. PSO-based path planners were developed in \cite{kathen2021informative} and \cite{gul2021meta}. In \cite{kathen2021informative}, the authors developed an informative path planner by combining the PSO components with the data obtained from the estimated water quality model generated by the GP. The authors compared the developed algorithm with other PSO-based algorithms in \cite{carolina2022comparison} and \cite{jara2022ola}, where in \cite{carolina2022comparison}, the focus of the algorithm is to explore the surface of the water resource and obtain the most reliable estimated model of the lake, and in \cite{jara2022ola} the focus is to exploit the area where the highest pollution peak of Ypacarai lake is located. In \cite{gul2021meta}, the authors propose a path planner based on PSO and Grey Wolf Optimization (GWO). The results show that the path planner proposed by the authors generates efficient paths, these paths being short and avoiding obstacles.

\subsection{Multi-modal multi-objective problems}

To overcome the limitations of the classical PSO for multimodal problems, different methods and strategies were implemented such as mutation mechanism, dynamic parameters, neighborhood techniques, among others. One of the methods studied by researchers to improve the PSO is the mutation method. The main idea is to improve the classical PSO by including mutation operators to preserve population diversity and overcome the problem of premature convergence. In \cite{khan2021multimodal}, the authors proposed a dynamic inertia weight to obtain an efficient balance between exploration and exploitation, and a mutation mechanism to diversify particles in the optimization process. In \cite{wang2013particle}, the authors employ three mutation strategies, Gaussian, Cauchy, and Lévy, to perform mutation on the local best and global best. The best mutation is selected according to the selection ratio of each mutation operator. The results show that not all types of problems can be effectively solved by one type of mutation operator. In \cite{luo2020hybridizing}, the authors develop a niching strategy to ensure population diversity. The authors connect each particle with its better in its neighborhood and create a raw species. Then, the dominant raw species are combined to generate the final species of the population.

Multimodal problems can also be solved using neighborhood information. \cite{zhang2020dynamic} proposed a dynamic $\epsilon$-neighborhood selection mechanism. The size of the neighborhood changes dynamically for each particle. In addition, the authors developed four different position-update strategies in which the particle position is updated according to the particle position and the behavior in its neighborhood. The results show that the proposed algorithm solves multimodal problems. However, it shows that it does not effectively solve high-dimensional multimodal problems. In \cite{zhang2020modified}, the global learning strategy is replaced by a dynamic neighborhood-based learning strategy. The velocity of each particle is updated with its best position and the best position of its neighborhood. In addition, the authors include an offspring competition mechanism to improve the performance of the proposed algorithm.

In \cite{chang2015modified}, the author divided the particle swarm into several subpopulations based on the order of the particles. The main idea is to replace the best position of the swarm with the best position of the subpopulation. Then, the particle velocity is updated with its best position and the best position of the subpopulation. In \cite{chang2017multimodal}, the concept of a subpopulation was also applied. The author develops an improvement to the PSO by dividing the swarm into two large groups, where one group focuses on the search for the global maximum and the other group on the global minimum. These two large groups are then divided into several subgroups. As in \cite{chang2015modified}, the particle velocity is updated using the best position of the subpopulation instead of the best position of the whole population. In \cite{zhang2019cluster}, the authors introduce a new cluster-based PSO that includes a leader updating mechanism and a ring topology. For subpopulation evolution, the global best PSO is used, and to improve the information interaction between subpopulations, a local best PSO with ring topology is adopted, which improves exploration.

Several strategies have been developed to improve the performance of the classical PSO. A self-adaptive PSO is developed in \cite{zhang2013improved}. To avoid getting stuck in local optima, the two acceleration coefficients of the PSO are dynamically modified at each iteration taking into account the Euclidean distance between the current position of the particles and the best position of all particles (global best). In \cite{khan2018modified}, the authors also proposed dynamically updating the PSO coefficients, the inertia weight, and the stochastic acceleration or acceleration coefficients. In the search process, each particle has a different value of inertia weight, and the values of the stochastic accelerations are changed randomly. In \cite{song2017global}, the authors adopted the multimodal delayed information strategy to improve the PSO. The particle velocity is updated according to the previous velocity, the local and global best, and two new terms, the local and global delayed best. These terms correspond to the values of the previous stochastic iterations of the local and global best. The algorithm evaluates the evolution factor at each iteration and selects the indicated evolutionary states (convergence state, exploitation state, exploration state, and the state of jumping out). A memetic PSO is proposed in \cite{wang2012memetic}. The authors combine the local PSO model and the Local Search (LS) method. The local PSO is used for global exploration of the search space, and for local exploitation, two different LS operators in a cooperative fashion are proposed. In addition, a method to trigger the reinitialization of the converged species is integrated.

This work fills several gaps with respect to previous approaches. First, previous works that focused on improving PSO-based path planners based on GP do not avoid local maximum/minimum \cite{jara2022ola}\cite{kathen2021informative}\cite{carolina2022comparison}; and also, they are not intended for multimodal scenarios. These works achieved satisfactory results in terms of model estimation but considered simple scenarios that only contain a global maximum. It can be expected in real scenarios that multiple contamination peaks are present. Second, none of the previous studies focused on multiagent systems for water monitoring applied federated learning \cite{luis2021multiagent}\cite{peralta2021bayesian}. These works used centralized systems for developing the estimated contamination models. The research presented here is the first attempt at the application of federated learning for water monitoring through ASV. This simplifies the communication congestion between vehicles and the main server. Third, the proposed approach is a step forward in terms of model estimation with respect to the proposed techniques based on exploration and exploitation phases like epsilon greedy algorithm and \cite{arzamendia2019intelligent}. The proposed approach achieves estimated models with very low errors in terms of MSE. 

\section{Problem formulation}\label{sec:statement}

\begin{figure}[htpb]
    \centering
    \includegraphics[width=0.6\textwidth]{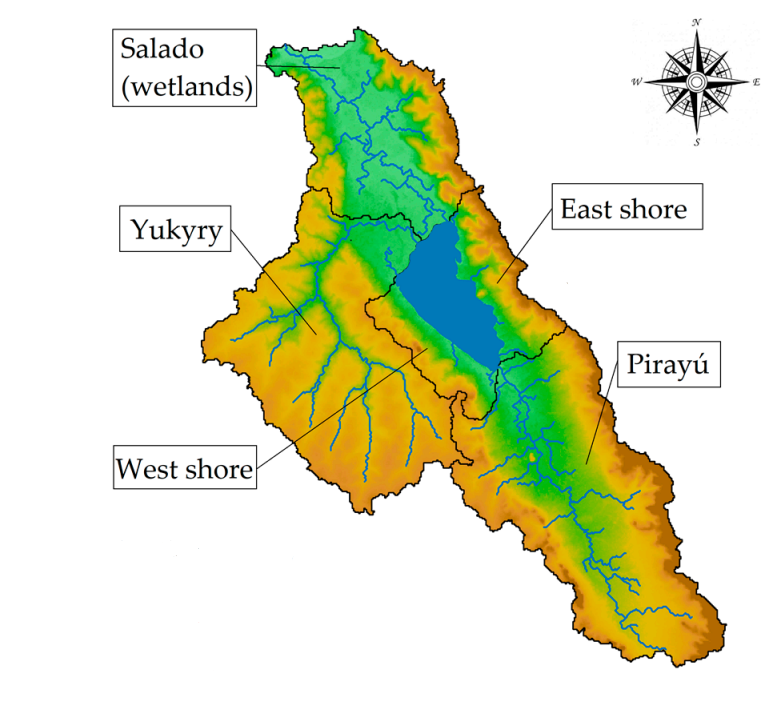}
    \caption{Inlets and outlet of the Ypacarai lake \cite{lopez2018eutrophication}}
    \label{fig:streams}
\end{figure}

As a case study, the situation of the Ypacarai lake is considered, Fig.~\ref{fig:streams}. The Ypacarai lake is the largest body of water in Paraguay (about $60 km^2$) and is surrounded by the cities of Luque, Areguá, Itauguá, Ypacaraí, and San Bernardino. The Ypacarai lake has two main inlets: 1) an inlet to the southeast, the Pirayú stream; and 2) an inlet to the northwest, the Yukyry stream. However, it also has other smaller inlets to the east and west of the lake. The lake has only one natural outlet, the Salado River, located north of the lake. The lake plays an important role in its environment; it is a source of water for the surrounding towns and farms, and the water is also used to irrigate the surrounding agricultural activities. During summer, the lake is used for recreational and tourist purposes \cite{arzamendia2019intelligent}. Therefore, the pollution of the lake affects not only nature, but also public health and the surrounding economy. The main sources of lake contamination are domestic and industrial wastewater from surrounding towns, as well as fertilizer residues that reach the lake from crops through rain or streams \cite{lopez2018eutrophication}. Because of this, constant monitoring of the water quality of the lake is carried out by governmental and nongovernmental institutions \cite{itaipu2018}\cite{itaipu2021}. The monitoring method applied is the traditional method, in which specialists take water samples at specific points around the lake and streams and analyze these samples in laboratories. Monitoring using ASV will allow not only to sample points around the lake, but also points all over the lake surface, allowing to obtain a more complete status of the water quality, to perform monitoring in less time and in a more efficient way.

In the proposed monitoring system, the ASVs are equipped with sensors $S$. These sensors take water quality measurements. For simulations, a real model of the water resource is considered, and this model is represented by the function $f(\bb{x})$, where $\mathbf{x}$ is the coordinate $(x, y)$ of the water resource. The $n$ measurements that are taken by the ASV sensors are stored in the vector $\mathbf{s} = \{ s_k  \; | \; k = 1,\: 2,\: \dots\: ,\: n\}$, where $k$ represents the number of measurements taken. In addition to the water quality data, the coordinates where the samples have been taken are also stored in the vector $\mathbf{q} = \{ q_k  \; | \; k = 1,\: 2,\: \dots\: ,\: n\}$. The combined data of water quality and coordinate measurements are stored in a vector $D = \{ (q_k, s_k) \; | \; k = 1,\: 2,\: \dots\: ,\: N\}$, representing the data taken during the monitoring of the water resource by the ASV fleet. For each $D_k$, the input is the position $q_k$ of the ASV and the output is the water quality data $s_k$ taken by the sensor:

\begin{equation}
    s_k = f(q_k)
\end{equation}

Applying the regression model, Eq.~\ref{eq:regression}, and having the necessary amount of data $D$, a real water quality model $f(\bb{x})$ of the Ypacarai lake can be obtained.

\begin{equation}
    y \approx f(\mathbf{x})\label{eq:regression}
\end{equation}

\begin{figure}[H]
    \centering
    \includegraphics[width=0.35\textwidth]{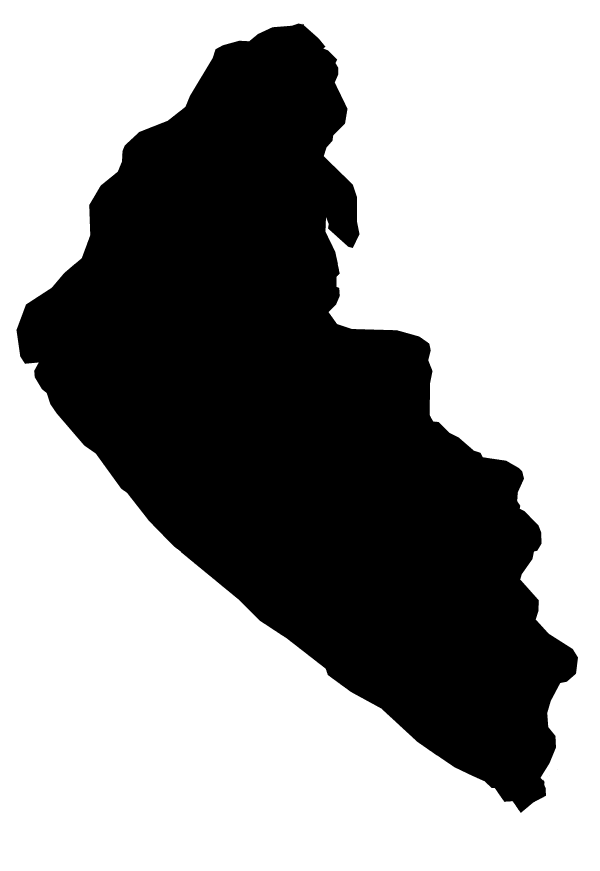}
    \caption{Model of the occupancy grid of the Ypacarai lake}
    \label{fig:gridmap}
\end{figure}

The following assumptions were made for the monitoring system:

\textbf{Assumption 1.} The monitoring system is performed in the Ypacarai lake, Fig.~\ref{fig:gridmap}. The matrix $\mathcal{N}$ represents the model space. The model (matrix $\mathcal{N}$) has a dimension of $m \times n$ and each element of the model $\mathcal{N}_{i,j}$ has a value that represents the state of the grid. The ASVs are allowed to travel to grids that have values equal to 1. Otherwise, the ASVs are not allowed to go to grids with a value equal to 0. This can occur because it is either land, or a prohibited zone, or an obstacle, among others. In the simulations, the values of the elements $\mathcal{N}_{i,j}$ are scaled, each grid measures 100 m x 100 m.

\textbf{Assumption 2.} The sensors are error-free. Therefore, the measurements taken by all sensors, water quality sensors, and GPS are accurate.

\textbf{Assumption 3.} The movements of the ASVs are error-free. Similarly, obstacles and collisions are not taken into account. The maximum speed at which the ASVs can travel is 2 m/s.

\textbf{Assumption 4.} In terms of battery power, the vehicle has enough autonomy to cover 30,000 meters.

\section{PSO-based Path planners} \label{sec:psobased}

This section describes different versions of PSO-based path planners for water monitoring with ASV.

\subsection{Classic Particle Swarm Optimization (PSO)}

The PSO algorithm is a population-based optimization technique developed by \cite{kennedy1995particle}. The concept of the algorithm was inspired by the social behavior of a school of fish and flocks of birds. The potential solutions of the optimization problem are called particles or individuals, and a group of particles is called swarm or population. The population of the PSO is initialized randomly. The particles travel in a multidimensional search space, sharing information with each other to expand the search area. Then, each particle follows a different path to reach the target. This allows more space to be covered to find the global optimum. The movement of the particles is obtained as a function of three components, the control parameter, the self-cognition component, and the social cognition component. Self-cognition or the local best component is the best particle position ($\bb{pbest}$) so far, and the social cognition or the global best component is the best position of the population ($\bb{gbest}$) so far. Each particle has a velocity $\bb{v}^{t}_p$ and position $\bb{x}^{t}_p$ that are updated according to Eq.~\ref{psovel} and Eq.~\ref{psopo}, respectively. 

\begin{subequations}
    \begin{align}
        \label{psovel}\bb{v}^{t+1}_p & = w \bb{v}^t_p + c_1r^t_1 \left[ \bb{pbest}^t_p - \bb{x}^t_p \right] + c_2 r^t_2 \left[ \bb{gbest}^t - \bb{x}^t_p \right]\\
        \label{psopo}\bb{x}^{t+1}_p & = \bb{x}^t_p + \bb{v}^{t+1}_p
    \end{align}
\end{subequations}

The term $w$ is the inertia weight or the control parameter. The best position of the particle $p$ at the $t$th iteration is represented by $\bb{pbest}^{t}_p$. Similarly, the best position of the population in the $t$th iteration is represented by $\bb{gbest}^{t}$. $c_1$ represents the cognitive constant and $c_2$ refers to the social constant. These two constants are also known as acceleration coefficients and describe the importance of the exploitation (local best) and exploration (global best) of the search space. $r_1$ and $r_2$ are random values between 0 and 1.

\subsection{Enhanced GP-based PSO}\label{sec:enhanced}
 
This approach is an improved version of the PSO path planner that employs the Gaussian process (GP) for surrogate modeling \cite{kathen2021informative}. The main idea is that the classical PSO algorithm is enhanced by adding the mean and uncertainty components given by the GP. Therefore, the vehicle will move guided by four components, best local, best global, GP mean, GP uncertainty. The main components of the Enhanced GP-based PSO are described as follows. 

\subsubsection{Gaussian Process}

The Gaussian process is a stochastic process based on Bayesian inference \cite{rasmussen2003gaussian}. The input to the GP consists of a random data set. The GP is completely defined by two components: the covariance function or kernel and the mean function \cite{rasmussen2003gaussian}. Nevertheless, the mean function is usually zero. For this monitoring system, the mean function is zero, so the component that defines the GP behavior in this algorithm is the kernel function. The kernel function specifies the smoothness, variability, and shape of the model of the water quality parameters of the water resource. According to the results obtained in \cite{peralta2021bayesian}, the Radial Basis Function (RBF) has the best behavior for water resources. 

To update the Gaussian Process Regression (GPR), the input data $D$ is conditioned and marginalized. Eq.\ref{eq:musigma} is used to calculate the unknown response $\mu(\mathbf{x_*})$ (mean value), $\sigma(\mathbf{x_*})$ (standard deviation) of the GPR $\hat{f}(\mathbf{x})_*$.

\begin{subequations}
    \begin{eqnarray}
        \mu_{\hat{f}(\mathbf{x}_i)_*\mid D} &=& K_*^T K^{-1} f(\mathbf{x})\\
        \sigma_{\hat{f}(\mathbf{x}_i)_*\mid D} &=& K_{**} - K_*^T K^{-1}K_*
    \end{eqnarray}
    \label{eq:musigma}
\end{subequations}

The terms $K$, $K_{**}$ and $K_*$ are data from the fitted kernel. These variables include covariances between known data $k(\mathbf{x}, \mathbf{x})$, unknown data $k(\mathbf{x}*, \mathbf{x}*)$, and covariances between both, the known and unknown data $k(\mathbf{x}, \mathbf{x}*)$.

\begin{gather}
K =
    \begin{bmatrix}
        K & K_* \\
        K_*^T & K_{**} 
    \end{bmatrix}   
    =
    \begin{bmatrix}
        k(\mathbf{x}, \mathbf{x}) & k(\mathbf{x}, \mathbf{x}*) \\
        k(\mathbf{x}*, \mathbf{x}) & k(\mathbf{x}*, \mathbf{x}*)
    \end{bmatrix}
\end{gather}

\subsubsection{Path Planner}

The Enhanced GP-based PSO is a path planner based on the PSO and GP. It combines the PSO components, the local best $\bb{pbest}$ and the global best $\bb{gbest}$, with the data obtained from the model estimated by the GP, the uncertainty $\sigma$ and the mean $\mu$, the latter serving the function of water resource contamination data. The algorithm takes water samples from the water resource through the water quality sensors and uses these samples to update the GP. Once the GP is adjusted, the path planner calculates the maximum value of the uncertainty and the mean to update the velocity and position of the ASVs using Eq.\ref{egpso}.

\begin{subequations}
    \begin{align}
    \begin{split}
    \label{egpsovel}\bb{v}^{t+1}_p&=w\bb{v}^{t}_p+{c}_{1}{r}^{t}_{1}[\bb{pbest}^{t}_p-\bb{x}^{t}_p]+{c}_{2}{r}^{t}_{2}[\bb{gbest}^{t}-\bb{x}^{t}_p] +{c}_{3}{r}^{t}_{3}[\bb{max\_un}^{t}-\bb{x}^{t}_p] + \\
                     &\quad{c}_{4}{r}^{t}_{4}[\bb{max\_con}^{t}-\bb{x}^{t}_p]
    \end{split}
    \\
    \label{egpsopo}\bb{x}^{t+1}_p & = \bb{x}^t_p + \bb{v}^{t+1}_p
    \end{align}
    \label{egpso}
\end{subequations}

The velocity $\textbf{v}^{t+1}$ is obtained from the sum of the velocity $\textbf{v}^{t}$ plus the data of the surrogate model and the components of the PSO. The term $w$ is the weighting of inertia. The data of the surrogate model are represented by $\textbf{max\_un}^t$ and $\textbf{max\_con}^t$. The $\textbf{max\_un}^t$ term is the coordinate where the maximum model uncertainty value $max\sigma^t$ is found. On the other hand, the $\textbf{max\_con}^t$ term is the coordinate where the maximum model mean value $max\mu^t$ is found, which refers to the maximum contamination of the water resource. The terms $c_1$, $c_2$, $c_3$ and $c_4$ are acceleration coefficients that determine the importance of each term. The higher the value, the greater the importance of the term. The coefficients $c_1$ and $c_4$ determine the importance of exploitation. The coefficients $c_2$ and $c_3$ determine the importance of exploration. Position $\textbf{x}^{t+1}$ is obtained by adding speed $\textbf{v}^{t+1}$ and position $\textbf{x}^{t}$. $r_3$ and $r_4$ are random values in the range [0, 1].

Water samples are not taken at each iteration of the algorithm, since this would take a large amount of data, which would consume a lot of time in the GP adjustment. As a condition for sampling, the distance between the sampling coordinates as described in Eq.~\ref{eq:s} is used \cite{peralta2021bayesian}. If the distance between the current position of the ASV $\mathbf{x}^t$ and the position where the last sample was taken $\mathbf{x}_{sample}$ is greater than the distance $l$, the water sample is taken.

\begin{equation}
    l = \lambda \times \ell^t\label{eq:s}
\end{equation}

The term $\lambda$ refers to the ratio of the length scale, and the term $\ell^t$ represents the posterior length scale of the GP.

\subsection{Enhanced GP-based PSO based on Epsilon Greedy method}
\label{sec:epsilon}

This is an improved version of PSO that combines exploration and exploitation phases using the epsilon greedy methodology, which is widely used in DRL \cite{yanes2021dimensional}. When the epsilon greedy method is applied to the Enhanced GP-based PSO, it allows the algorithm to change focus during the monitoring task. This is due to the epsilon function $\epsilon$, which allows the algorithm to randomly change the values of the acceleration coefficients ($c_1, c_2, c_3$ and $c_4$) of the Enhanced GP-based PSO. The epsilon function $\epsilon$ represents the probability of exploration.

The operation of the Enhanced GP-based PSO based on the epsilon greedy method is shown in Algorithm 1. First, the combination of acceleration coefficient values for exploration ($Explore$) and exploration ($Exploit$) must be selected. Once the algorithm is started, the probability that ASVs will engage in surface exploration is 95$\%$, since the value of epsilon $\epsilon$ is equal to 0.95. When ASVs reach a distance traveled $d\epsilon_0$, the epsilon value $\epsilon$ starts to decay $\Delta\epsilon$. The epsilon value $\epsilon$ decays until the ASVs reach a distance traveled equal to $d\epsilon_f$. When this distance is reached, the epsilon value $\epsilon$ remains constant again. However, the value at which it remains constant is 0.05, which means that the probability of exploring the surface of the water resource is 5$\%$, giving more possibility to exploit the contamination zones. The condition that determines whether the algorithm focuses on exploration or exploitation is the relationship between the epsilon $\epsilon$ and the random number $val$, which is generated at each iteration of the algorithm. If the value of epsilon $\epsilon$ is greater than the value of $val$, the ASVs focus on exploring the surface. On the contrary, if the epsilon value $\epsilon$ is less than the $val$ value, the ASVs focus on exploiting the contamination zones. After determining the values of the coefficients, the path planner behaves like the Enhanced GP-based PSO.

\begin{figure}[htpb]
    \begin{algorithm}[H]\label{pseudo}
    \caption{Enhanced GP-based PSO based on Epsilon Greedy method pseudo-code}
    $\bb{x}^0_p \ \gets$ Initialize PSO\tcp*[l]{where $\bb{x}^0$ is the initial position of the ASV $p$} 
    \While{not done}
    {
      \If{${dist}_\text{total} \leq d\epsilon_0$}
        {$\epsilon^t \gets 0.95$}
      \ElseIf{${dist}_\text{total} \geq d\epsilon_f$}
        {$\epsilon^t \gets 0.05$}
      \Else
        {$\epsilon^t \gets \epsilon^{t-1}-\Delta\epsilon$}
      $val\gets random()$\;
      \If{$\epsilon^t \geq val$}
        {$c_1, c_2, c_3, c_4 \gets$ Explore}
      \Else
        {$c_1, c_2, c_3, c_4 \gets$ Exploit}
      $\bb{pbest}_p,\ \bb{gbest} \ \gets$ Evaluate fitness function\;
      $dist \gets \bb{x}^t - \ \bb{x}_{sample} \ \gets$ Calculate distance\;
      \If{$dist \geq l$}
      {
        $s \ \gets$ Take water sample\; 
        ${\sigma}^t,\ {\mu}^t \ \gets$ Adjust GP\;
        $max\sigma^t,\ max\mu^t \ \gets$ Find maximum values\;
        $\bb{max\_un}^{t}, \ \bb{max\_con}^{t} \ \gets$ Obtain coordinates of the maximum values\;
      }
      $\bb{v}^{t+1}_p,\ \bb{x}^{t+1}_p \ \gets$ Update speed and position of the ASVs\;
    }
    \end{algorithm}
\end{figure}

\section{Proposed system monitoring: AquaFeL-PSO}\label{sec:proposed}

The main objective of this section is the development of an informative path planner based on a multimodal PSO, GP, and FL paradigm. The new monitoring system uses as a basis the Enhanced GP-based PSO (Sect.~\ref{sec:enhanced}). Improvements are added to increase accuracy and precision in detecting and monitoring areas of high contamination. The new improvement is based on dividing the monitoring system into two phases, exploration and exploitation. Contrary to the epsilon greedy method (Sec.~\ref{sec:epsilon}), in this new system, the phases have controlled periods.

\begin{figure}[htpb]
    \centering
    \includegraphics[width=0.3\textwidth]{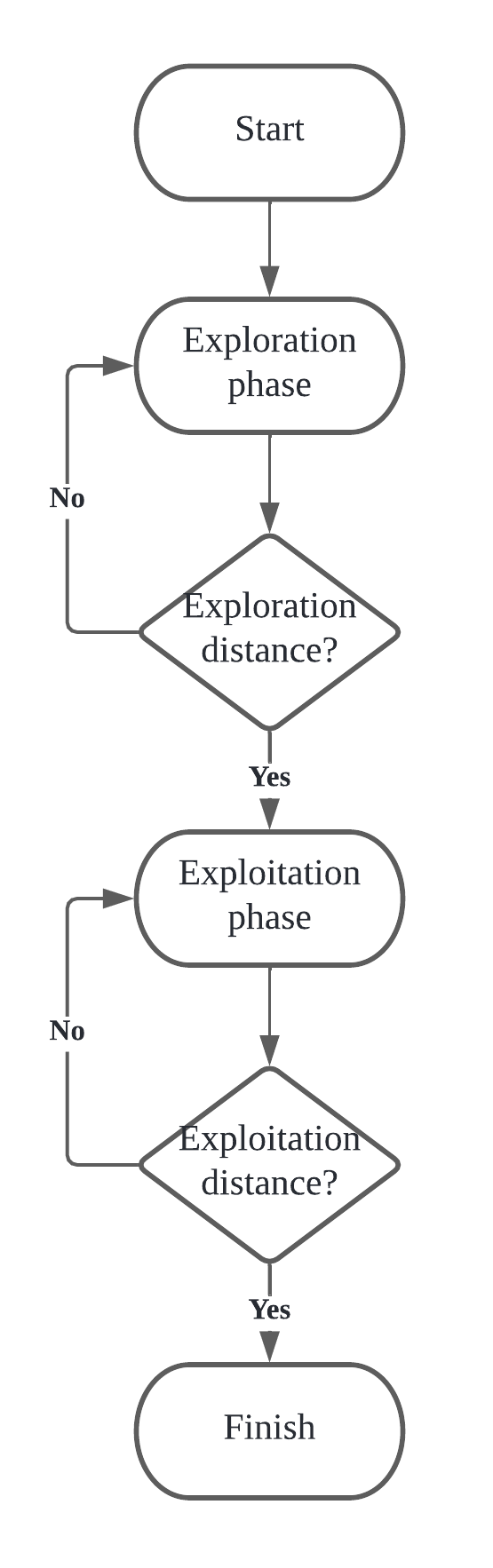}
    \caption{Flowchart of the proposed monitoring system}
    \label{flow}
\end{figure}

Fig.~\ref{flow} shows the flowchart of the the proposed monitoring system, which starts with focusing on the exploration of the surface water resource. The objective of this first phase is to obtain a reliable water quality model to determine the areas of highest contamination of the water resource, which are called action zones. Then, when the distance traveled by the ASVs reaches a certain distance, called the exploration distance, the estimated water quality model is obtained with the samples taken in the exploration phase. Once the model is obtained, the system moves on to the exploitation phase. The exploitation phase consists of taking water quality samples in areas where the highest level of contamination is detected. In this second phase, the surface of the water resource is divided into action zones and the ASVs are assigned to these zones. Decision and assignment of ASVs are done only once. The ASVs are assigned according to the distance of the ASV from the center of the action zone. Once the ASVs are assigned, the vehicles go directly to the corresponding action zone and exploit it to obtain a more accurate model. Each action zone has its own estimated model, having as starting point the model obtained in the exploration phase. The vehicles finish the monitoring task when they reach a distance traveled equal to the exploitation distance. Finally, the estimated models are merged to obtain a single water quality model of the water resource. The values of exploration distance and exploitation distance are parameters that are set at the start of the algorithm.

\subsection{Exploration phase}

\begin{figure}[H]
    \centering
    \includegraphics[width=0.5\textwidth]{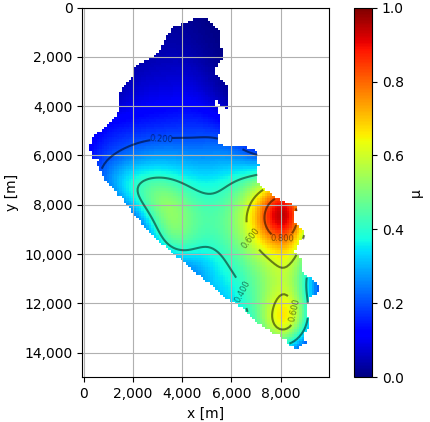}
    \caption{An example of the water quality model obtained in the exploration phase}
    \label{explorphase}
\end{figure}

In this first phase, the ASVs focus on covering and sampling as much of the water resource as possible. The objective is to obtain a first estimated model of the water quality of the water resource so that later, in the exploitation phase, this model can be used to determine the action zones. An example of the estimated water quality model of the water resources obtained in this phase can be seen in Fig.~\ref{explorphase}. At this stage, a centralized learning technique is used, all samples taken by the vehicles are sent to the central server, and a single estimated model is generated. This phase ends when the vehicles have traveled a certain distance. Once this distance is reached, the monitoring system switches the focus to exploitation.

Since the proposed monitoring system is based on the Enhanced GP-based PSO, the velocity and position update is based on Eq.~\ref{egpsovel}. Since in this first phase, the objective of the monitoring task is to obtain a good first estimate of the water quality model of the water resource. Therefore, the acceleration coefficients that remain active are the local best ($c_1$) and the surrogate model uncertainty ($c_3$) coefficients, since it has been demonstrated that with these coefficients the monitoring system focuses on exploring the surface of the water resource \cite{carolina2022comparison}. As a consequence, the new velocity equation for this phase is Eq.~\ref{phase1vel}. The position is updated using the same equation as the Enhanced GP-based PSO, Eq.~\ref{egpsopo}.

\begin{equation}
\label{phase1vel} \bb{v}^{t+1}_p=w\bb{v}^{t}_p+{c}_{1}{r}^{t}_{1}[\bb{pbest}^{t}_p-\bb{x}^{t}_p +{c}_{3}{r}^{t}_{3}[\bb{max\_un}^{t}-\bb{x}^{t}_p]
\end{equation}

\subsection{Exploitation phase}

The exploitation phase is divided into different stages. This subsection describes the stages of this phase.

\subsubsection{Action zones}

\begin{figure}[htpb]
    \centering
    \includegraphics[width=0.5\textwidth]{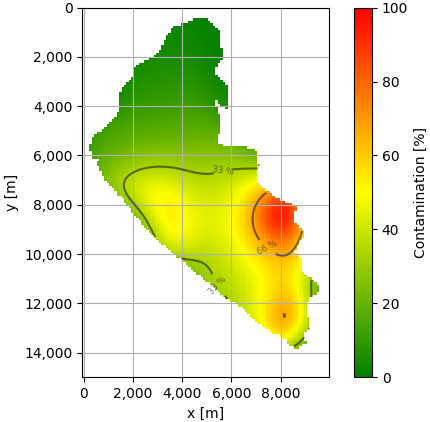}
    \caption{An example of the estimated model taking into account the percentage of contamination}
    \label{explorelevels}
\end{figure}

Using the estimated model generated in the exploration phase, areas with high contamination levels are found. The considered contamination levels are: acceptable 0 - 32$\%$ (green); warning 33$\%$ - 65$\%$ (yellow); and risk 66$\%$ - 100$\%$ (red). The percentage of contamination is determined from the maximum value of contamination found in the exploration phase, so it would be, as an example, 33$\%$ of the maximum value of contamination in the exploration phase. Fig.~\ref{explorelevels} represents an example of the estimated model obtained in the exploration phase taking into account the percentage of contamination. In this work, circular action zones are considered. The idea is that the peaks of the contamination values of the estimated model obtained by the exploration phase represent the centers of the action zones.

\begin{itemize}
    \item \textbf{Area of action zones:}  To obtain the action zones, a radius of action $rad$ is calculated that determines the area covered by vehicles during the exploitation phase. The area of action zones should adapt to length of water resources, and also, the number of vehicles available for the monitoring mission. Thus, if more vehicles can be used, the resolution of the action zones can be increased by decreasing the area of the action zone. Therefore, the proposed action radius is shown in Eq.~\ref{eq:rad_action}, the action radius is a function of the shortest length of the lake $length$ divided by the number of vehicles $n_{ASVs}$ in the ASVs fleet.
    
    \begin{equation}
    rad = length/n_{ASVs}\label{eq:rad_action}
    \end{equation}
    
    The areas of the action zones do not overlap, since the coordinates that already belong to a zone are discarded to obtain the next action zone.
    
    \item \textbf{Number of action zones:} To determine the number of action zones, first a contamination threshold has to be selected; if the contamination value at a certain coordinate is higher than the contamination threshold, that coordinate is considered as a possible action zone center. The selected contamination threshold is 33$\%$ of the maximum value of the contamination of the exploration phase, since it is considered that below such threshold, water is in optimal condition. Among all coordinates that meet this requirement, the coordinates of the maximum value are considered the center of the first action zone. Then, the procedure is recursive by eliminating the area within the first action zone, and consequently, determining the center of the second action zone among the rest of the water resources that are above the 33$\%$ of the maximum value of contamination of the exploration phase. The procedure finishes when the maximum number of action zones is determined. The maximum number of action zones is the number of vehicles in the fleet. Notice that between the centers of two action zones, there will never be a distance smaller than the action radius because the center of the action zone is obtained considering the coordinates outside the circumference. As a consequence of the proposed procedure for determining the action zones, two cases can occur: 1) there can be fewer action zones than vehicles, this is because there are fewer pollution peaks than ASVs; 2) there can be the same number of action zones as vehicles, this is because there are the same number of pollution peaks as vehicles. In this work, it is not considered to have action zones than vehicles, since the number of pollution peaks is limited by the number of vehicles.
\end{itemize}

\subsubsection{Resource allocation}

Once the action zones have been obtained, the vehicles are assigned to the action zones with the objective that ASVs exploit the zone. The ASVs are assigned according to the distance between the vehicle and the center of the action zone. The ASV that is closest to the action zone is the one that takes advantage of that zone. In case there are more ASVs than action zones, more than one vehicle is assigned in the action zones. The priority of the action zone is the condition to determine the assignment of the vehicles in this case. The priority of the action zones is determined by the peak contamination value of the action zone. Thus, the highest priority would be given to the action zone that has the highest peak contamination of the model estimated in the exploration phase. This parameter is used to assign vehicles to action zones in case there are more vehicles than action zones. The higher the priority of the action zone, the higher the probability that there will be more ASVs exploiting the area. The maximum value of the priority $max\_prt$ depends on the number of vehicles in the fleet and is obtained by applying Eq.~\ref{eq:priority}:
    
\begin{equation}
max\_prt = n_{ASVs} * 10 + 10\label{eq:priority}
\end{equation}

\noindent where $n_{ASVs}$ is the number of vehicles. Each time a vehicle is assigned to an action zone, the priority decreases by 10 points.

\subsubsection{Exploitation of action zones}

After obtaining the action zones and assigning the ASVs, the actual exploitation task is conducted. By dividing the fleet into different action zones, the swarm is divided into subpopulations or multiple swarms. Each subpopulation focuses on the exploitation of its action zone, hence the maximum contamination and maximum uncertainty data are different for each subpopulation, as well as the global best. To calculate the maximum value of each action zone, only the data of the coordinates that are in that action zone are taken into account, discarding the other coordinates of the map. In this way, the exploitation of this action zone is not affected by the other action zones. Moreover, each subpopulation calculates a different global best; therefore, the global best of a subpopulation is the best position among the vehicles that are in the same action zone. If there is only one vehicle in the action zone, the global best is equal to the local best. To end with the monitoring, the vehicles must travel a certain distance, which is the exploitation distance. As in the exploration phase, in the exploitation phase, Eq.~\ref{egpso} is used as the basis for updating velocity and position. In this phase, the acceleration coefficients that remain active are the local best ($c_1$), global best ($c_2$) and contamination ($c_4$) coefficients, with higher values for the local best and contamination coefficients. It has been demonstrated in \cite{jara2022ola}, that with this combination, the algorithm exploits the zones through which the ASV passes and the zone with the highest combination. Therefore, the velocity update equation in the exploitation phase is left as Eq.~\ref{phase2vel}, and the position update is performed by applying Eq.~\ref{egpsopo}.

\begin{equation}
\label{phase2vel}\bb{v}^{t+1}_p=w\bb{v}^{t}_p+{c}_{1}{r}^{t}_{1}[\bb{pbest}^{t}_p-\bb{x}^{t}_p]+{c}_{2}{r}^{t}_{2}[\bb{gbest}^{t}-\bb{x}^{t}_p] +{c}_{4}{r}^{t}_{4}[\bb{max\_con}^{t}-\bb{x}^{t}_p]
\end{equation}

\subsubsection{Federated/distributed Learning}

To obtain the estimated water quality model of the water resource, the concept of a machine learning technique, Federated Learning (FL), is applied. The FL technique is a new concept that was introduced by McMahan et al., with the aim of updating language models in cell phones \cite{mcmahan2016federated}\cite{mcmahan2017communication}\cite{konevcny2016federated}\cite{konevcny2016federatedb}. The idea of the FL is to generate an ensemble ML model \cite{yang2019federated}. The data that are used to generate this model are scattered in multiple sites. These data are trained by the local servers or nodes, and the information that is shared with the central server is the result of the training. In this way, the security of the data of each node and the privacy of each node are preserved. The final result obtained in the central server is very similar to the final model generated by a system with a centralized learning technique \cite{yang2019federated}. In addition to promoting the privacy and security of the framework, nodes can train ML models in a collaborative manner, allowing better results to be generated \cite{yang2019federated}\cite{chen2020wireless}. Among the drawbacks of the FL are: the communication between nodes and the central server can be unstable and slow; and the more nodes there are, the more likely the system will become unpredictable and unstable \cite{yang2019federated}\cite{chen2020wireless}.

For a better understanding, reference is made to the definition provided by \cite{yang2019federated}. Consider $\mathcal{L}$ nodes, with the vector of nodes being equal to $\mathbf{F} = \{ \mathcal{F}_k  \; | \; k = 1,\: 2,\: \dots\: ,\: \mathcal{L}\}$. These nodes will train ML models using their respective data $\mathbf{W} = \{ \mathcal{D}_k  \; | \; k = 1,\: 2,\: \dots\: ,\: \mathcal{L}\}$. In the case of a central server, the data of the node $\mathcal{D} = \mathcal{D}_1 \cup \mathcal{D}_2 \cup ... \cup \mathcal{D}_\mathcal{L}$ would be joined together to obtain a model $\mathcal{M}_{CEN}$. However, with FL, each node trains its own model $\mathcal{M}_{FED}$ without exposing the privacy of its data. Moreover, the accuracy of the model generated with the FL $\mathcal{M}_{FED}$, being the accuracy $\mathcal{V}_{FED}$, is quite close to the model generated with the central server $\mathcal{M}_{CEN}$, $\mathcal{V}_{CEN}$. That is, considering $\delta$ a nonnegative real number, if 

\begin{equation}
    |\mathcal{V}_{FED} - \mathcal{V}_{CEN}| < \delta\label{eq:fed}
\end{equation}

Then, the federated learning algorithm suffers a loss of accuracy $\delta$.

\begin{figure}[htpb]
    \centering
    \includegraphics[width=\textwidth]{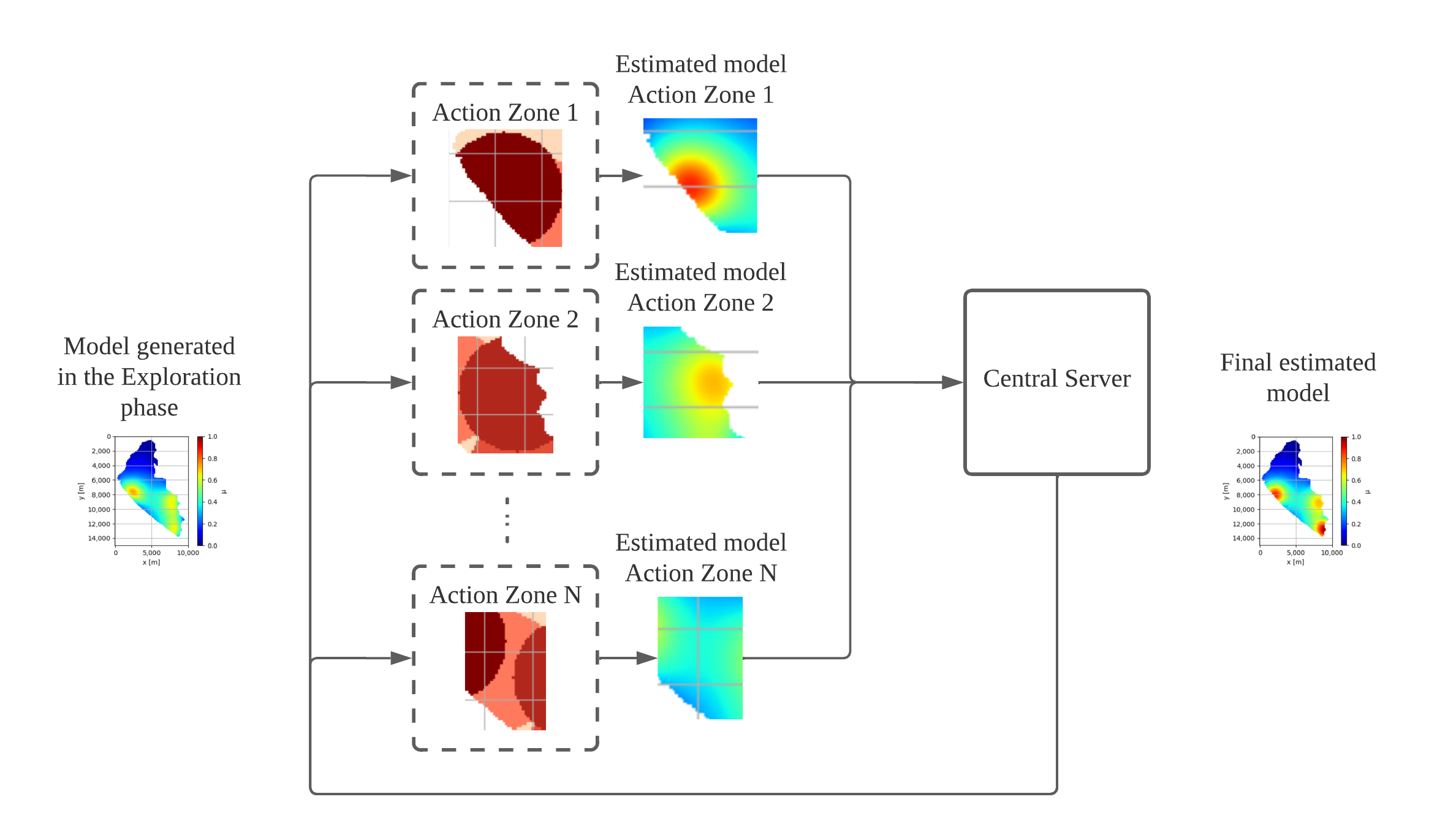}
    \caption{Process based on Federated Learning}
    \label{federated}
\end{figure}

Once the definition of FL has been clarified, the FL technique in the proposed monitoring system is explained. Fig.~\ref{federated} shows the process that is fulfilled based on the FL. For each action zone, a different estimated model is generated, taking into account the previous data from the exploration phase and the new data collected by the vehicles within that action zone. Therefore, the generation of the models is performed at each node or action zone. For the final water resource model, the results obtained at each node are replaced by the model generated in the exploration phase, at the central server. The data to be replaced are the mean and uncertainty values of the model obtained in the corresponding action zone. 

\section{Simulation Results}\label{sec:results}

This section contains the performance evaluation of the proposed monitoring system. First, the main simulation parameters used are described, including the Ypacarai lake target scenario, the performance metrics used for evaluation, and the values of the coefficients used in the proposed algorithm. Second, we conduct several analysis to validate the proposed approach in terms of the scalability and duration of the exploration and exploitation phases. Finally, the proposed approach is compared with other counterparts available in the literature.

\subsection{Simulation settings}

The code was developed in Python 3.8 using the Scikit-learn\footnote{https://scikit-learn.org/stable/ (accessed on 21 June 2022)}, DEAP\footnote{https://deap.readthedocs.io/en/master/ (accessed on 21 June 2022)} and Bayesian Optimization\footnote{https://github.com/fmfn/BayesianOptimization (accessed on 21 June 2022)} libraries. The code is available online at Github\footnote{https://github.com/MicaelaTenKathen/AquaFeL-PSO.git (accessed on 28 June 2022)}. The simulations were carried out on a laptop computer with 8GB RAM, Intel i5 1.60 GHz processor.

\subsubsection{Simulation scenario: Ypacarai Lake}

Simulations are carried out using the Ypacarai lake as the ground truth. Each element of the map matrix has a dimension of 100 x 100 meters. The distribution map of water quality parameters is obtained through a multimodal, multidimensional, continuous, and deterministic benchmark function, the Shekel function, Eq.~\ref{eq:shekel}

\begin{equation}
    f_\text{Shekel}(\mathbf{x}) = \sum_{i = 1}^{M} \frac{1}{c_{i} + \sum_{j = 1}^{N} (x_{j} - a_{ij})^2 }\label{eq:shekel}
\end{equation}

\begin{figure}[H]
\centering
     \begin{subfigure}{0.49\textwidth}
     \centering
         \includegraphics[width=\textwidth]{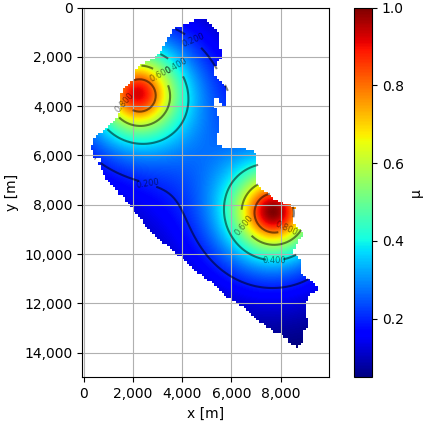}
         \caption{}
         \label{levels}
     \end{subfigure}
     \begin{subfigure}{0.49\textwidth}
     \centering
         \includegraphics[width=\textwidth]{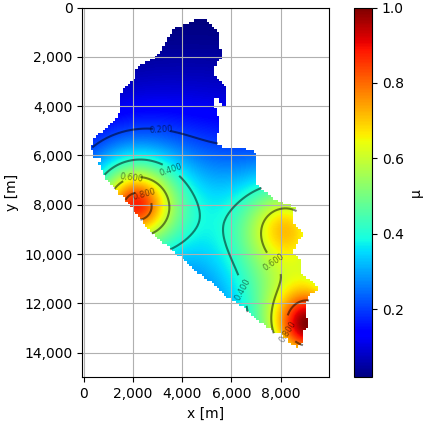}
         \caption{}
         \label{levels2}
     \end{subfigure}
     \caption{Examples of the actual water quality model of the water resource}
     \label{ground}
\end{figure}

The Shekel function, being a multimodal function, allows having several maximum points. These maximum points represent the highest levels of contamination of the Ypacarai lake. Some examples of the Shekel function applied to the Ypacarai lake scenario are shown in Fig.~\ref{ground}. The Shekel function is made up of two elements, the positions where the maxima are found, $a_{ij}$, and the inverse of the significant value of the maxima of the function, $c_{i}$. The element $a_{ij}$ is in the matrix $A$ and the element $c_{i}$ belongs to the matrix $C$. The matrix $A$ has a size of $M \times N$, where $M$ is the number of maximum points of the function and $N$ is the dimension of the space. On the other hand, the matrix $C$ has a size of $M \times 1$. For the simulations to be performed, the number of peaks, $M$, varies between 2 and the number of vehicles, and the dimension of the space $N$ will be 2, $x$ and $y$. The values of the matrix $C$ will be obtained randomly.

\paragraph{Normalization of data} 

\sloppy Because the values of the matrix $C$, in Eq.~\ref{eq:shekel}, are obtained randomly, the range of values of the water quality parameters in each ground truth varies. To standardize the range of values between 0 and 1, normalization is applied to the Shekel function data, Eq.~\ref{norm}. 

\begin{equation}
    \label{norm}f_\text{Normalized}(\mathbf{x}) = \frac{f_\text{Shekel}(\mathbf{x})-f_\text{min\_Shekel}(\mathbf{x})}{f_\text{max\_Shekel}(\mathbf{x})-f_\text{min\_Shekel}(\mathbf{x})}
\end{equation}

The term $f_{Shekel}(\bb{x})$ represents the current value of the function, the maximum value of the Shekel function is represented by $f_{max\_Shekel}(\bb{x})$, and the minimum value of the Shekel function is represented by the term $f_{min\_Shekel}(\bb{x})$.

\subsubsection{Performance metrics}

The results that are compared between the path planners are the results obtained with the estimated model of the whole lake, the estimated model in the action zones, and the value obtained for the peaks of each action zone. For this purpose, the metrics used are the mean square error (MSE) for the estimated models; and the error between two points, for the value of the peaks of each action zone. The MSE has been selected because this metric is often used to analyze the results of regression models. To calculate the MSE of the action zones and the error in the maximum peaks, the action zones of the ground truth were considered. In other words, the division of the map into action zones was applied to the ground truth, since the ground truth is considered as the real model of the water quality of the Ypacarai lake. From these real action zones, the maximum peaks of real contamination were obtained. Then, the MSE and error were calculated taking into consideration the coordinates of the real contamination peaks and real contamination zones.

For the first case of comparison, the objective is to exploit the action zone and obtain the most accurate possible model of the action zone, which represents a water resource contamination zone. For this purpose, Eq.~\ref{eq:msecon} is applied:

\begin{equation}
    \label{eq:msecon}\text{MSE}_\text{$action\_zone$}(f(\mathbf{x}), y) = \frac{1}{n_\text{$action\_points$}} \sum_{k=0}^{n_\text{$action\_points$} - 1} (f(\mathbf{x}_k) - y_k)^2
\end{equation}

\noindent where $f(\mathbf{x})$ represents the ground truth values in the action zones and $y$ represents the estimated model in the action zone.

For the second case of comparison, the objective is to detect the peaks of the action zones. To evaluate the performance of the path planners, Eq.~\ref{eq:error} is applied, where the error between the value of the contamination peaks detected in the ground truth $f(\mathbf{x}_{action\_zone\_peak})$ and the contamination peaks detected in the estimated model $y_{action\_zone\_peak}$ is calculated.

\begin{equation}
    \label{eq:error}\text{Error}_\text{$action\_zone\_peak$}(f(\mathbf{x}), y) =  |f(\mathbf{x}_{action\_zone\_peak}) - y_{action\_zone\_peak}|
\end{equation}

Finally, the objective is to obtain the estimated water quality model of the entire lake surface as accurately as possible. Due to this, using Eq.~\ref{eq:mse}, the MSE is calculated between the actual water quality model, $f(\mathbf{x})$, of the entire lake surface and the model estimated by the GP, $y$.

\begin{equation}
    \label{eq:mse}\text{MSE}_\text{$map$}(f(\mathbf{x}), y) = \frac{1}{n_\text{$map\_points$}} \sum_{k=0}^{n_\text{$map\_points$} - 1} (f(\mathbf{x}_k) - y_k)^2
\end{equation}

\subsubsection{Simulation parameters}\label{settings}


The maximum speed at which vehicles can reach is 2 meters per second. The value of the length scale of the GP is set to 10, taking into account the above mentioned by \cite{peralta2021}, the authors indicate that to have a good smoothness, the value of the length scale should be 10$\%$ of the value of the search map. Considering the works in \cite{peralta2021bayesian} and \cite{kathen2021informative}, the range for $\lambda$ to obtain a good performance of the monitoring system is [0.1, 0.5]. By setting the value of $\lambda$ to 0.1, the ASVs will take more samples. However, the execution time will be higher. On the contrary, if the value of $\lambda$ is 0.5, the monitoring system will take less time, on the other hand, the value of samples will be lower. To balance these two factors, the $\lambda$ value was set to 0.3, the middle value of the range. The values of the acceleration coefficients are different in the exploration and exploitation phase. In the exploration phase, the values of the coefficients obtained in \cite{carolina2022comparison} are used, and in the exploitation phase, the values of the coefficients obtained in \cite{jara2022ola} are used. This is because in these works studies were carried out in which the behavior of the algorithm in exploration and exploitation was analyzed. Data of the acceleration coefficients are shown in Table~\ref{coefphases}. These values are used for all fleet sizes.

\begin{table}[H]
    \caption{Values of the acceleration coefficients for exploration and exploitation phases}
    \centering
    \begin{tabular}{ccc}
    \hline
    Hyper-parameter	& Exploration phase& Exploitation phase\\
    \hline
    $c_1$    	& 2.0187 & 3.6845		\\\hline
    $c_2$   	& 0 & 1.5614		\\\hline
    $c_3$   	& 3.2697 & 0		\\\hline
    $c_4$   	& 0 & 3.6703  	\\
    \hline
    \end{tabular}
    \label{coefphases}
\end{table}

\subsection{Proposed approach evaluation}

In this subsection, different tests are carried out to evaluate the proposed monitoring system.

\subsubsection{Action zone scalability and resource allocation}

Firstly, we evaluate the scalability of the proposed approach related to the action zones. Fig.~\ref{fig:vehicles} shows some examples for different number of vehicles. The figures on the left show the models obtained in the exploration phase, and the figures on the right show the division of the water resource map into action zones depending on the number of ASVs in the fleet. This configuration allows that, by having more vehicles and smaller action zones, the polluted areas in the water resource can be divided into more zones and more vehicles will be exploiting different areas of the pollution zone, as shown in Fig.~\ref{8v}.

The Table~\ref{assig} shows the radius $rad$ and maximum priority $max\_prt$ parameters of the action zones for each number of vehicles $n_{ASVs}$. The results were obtained by applying Eq.~\ref{eq:rad_action} and Eq.~\ref{eq:priority} respectively. The length parameter $length$ of the Eq.~\ref{eq:rad_action}, in the Ypacarai lake scenario has a value of 10,000 meters.

\begin{table}[H]
    \centering
    \caption{Parameters of the action zones according to the number of ASVs in the fleet}
    \label{assig}
    \begin{tabular}{ccc}
    \hline
         Number of ASVs & $rad$ [m] & $max\_prt$ \\
         \hline
          2 & 5,000 & 30 \\\hline
          4 & 2,500 & 50\\\hline
          6 & 1,667 & 70\\\hline
          8 & 1,250 & 90\\\hline
    \end{tabular}
    
\end{table}

In Fig.~\ref{fig:vehicles}, in the figures on the right, the colors represent the priority of the action zones; the higher the priority, the higher the pollution in that action zone. The priority is used for the allocation of vehicles. If there are more vehicles than action zones, excess vehicles will be assigned to the zones with higher priority.

\begin{figure}[H]
\centering
     \begin{subfigure}{0.49\textwidth}
     \centering
         \includegraphics[width=\textwidth]{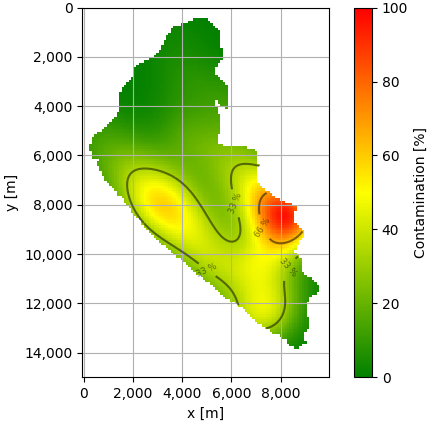}
         \caption{Estimated model obtained on exploration phase with 2 vehicles}
         \label{2vl}
     \end{subfigure}
     \begin{subfigure}{0.49\textwidth}
     \centering
         \includegraphics[width=\textwidth]{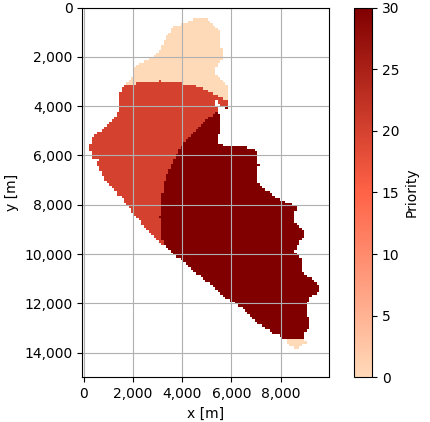}
         \caption{Action zones obtained with 2 vehicles}
         \label{2v}
     \end{subfigure}
     \begin{subfigure}{0.49\textwidth}
     \centering
         \includegraphics[width=\textwidth]{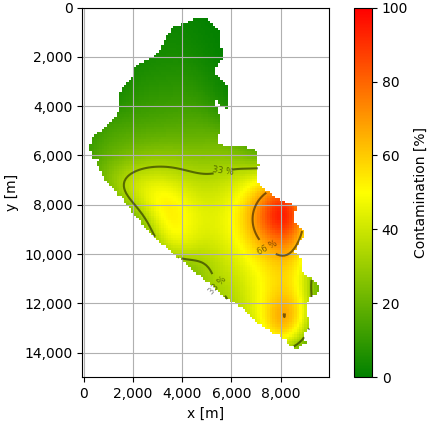}
         \caption{Estimated model obtained on exploration phase with 4 vehicles}
         \label{4vl}
     \end{subfigure}
     \begin{subfigure}{0.49\textwidth}
     \centering
         \includegraphics[width=\textwidth]{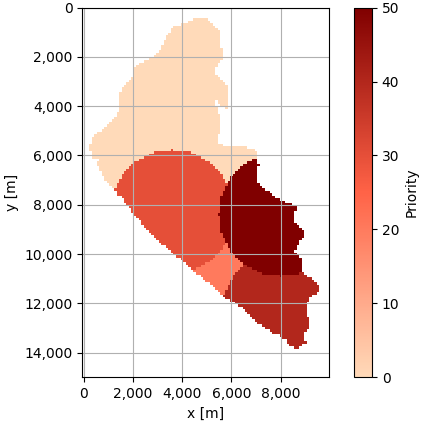}
         \caption{Action zones obtained with 4 vehicles}
         \label{4v}
     \end{subfigure}
\end{figure}

\begin{figure}[H]
\ContinuedFloat
    \centering
     \begin{subfigure}{0.49\textwidth}
     \centering
         \includegraphics[width=\textwidth]{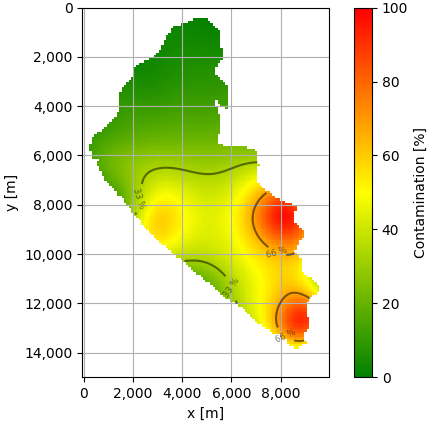}
         \caption{Estimated model obtained on exploration phase with 6 vehicles}
         \label{6vl}
     \end{subfigure}
     \begin{subfigure}{0.49\textwidth}
     \centering
         \includegraphics[width=\textwidth]{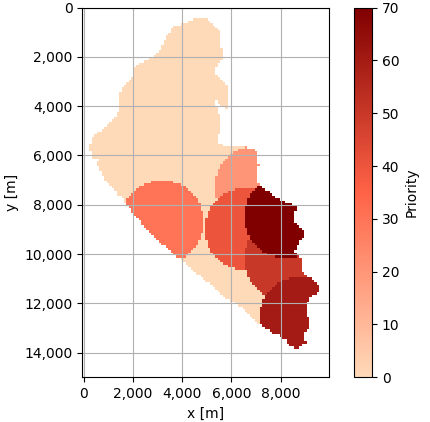}
         \caption{Action zones obtained with 6 vehicles}
         \label{6v}
     \end{subfigure}
     \begin{subfigure}{0.49\textwidth}
     \centering
         \includegraphics[width=\textwidth]{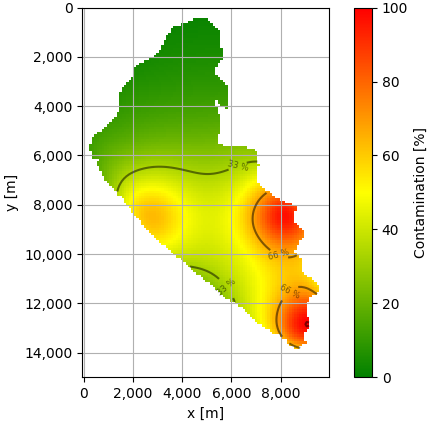}
         \caption{Estimated model obtained on exploration phase with 8 vehicles}
         \label{8vl}
     \end{subfigure}
     \begin{subfigure}{0.49\textwidth}
     \centering
         \includegraphics[width=\textwidth]{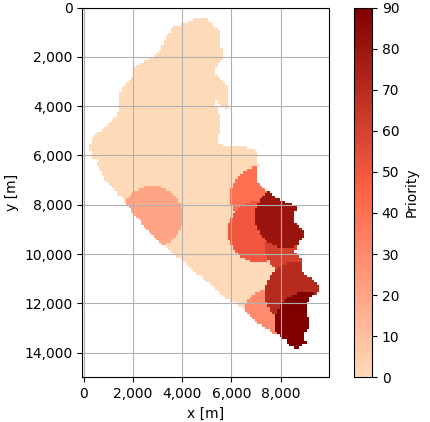}
         \caption{Action zones obtained with 8 vehicles}
         \label{8v}
     \end{subfigure}
    \caption{Examples of water resource action zones considering the number of ASVs. The figures on the left represent the estimated model obtained in the exploration phase in pollution percentage values, and the figures on the right represent the action zones obtained according to the number of vehicles in the fleet.} 
    \label{fig:vehicles}
\end{figure}

\subsubsection{Exploration vs Exploitation duration}

Before comparing the performance of the proposed approach with other path planners, the best combination for the duration of exploration and exploitation phases must be determined. For this purpose, tests were performed considering different distances for the exploration and exploitation phases. The maximum distance for these tests was 30,000 meters. However, the results are analyzed for a maximum distance of 20,000 meters and 30,000 meters. The tests were performed with two, four, six, and eight ASVs. The results obtained with two, six, and eight ASVs are shown in \ref{sec:sample:appendix}. The values simulated for the change from exploration phase to exploitation vary by 5,000 meters or 5km. In the tables, the measurement of km is used instead of meters.

\begin{figure}[H]
\centering
     \begin{subfigure}{0.49\textwidth}
     \centering
         \includegraphics[width=\textwidth]{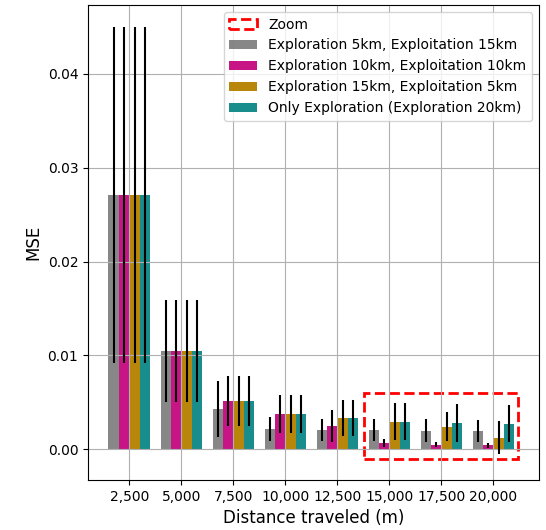}
         \caption{Mean of the MSE with 95$\%$ confidence interval of the cases studied for the duration of the exploration and exploitation phases}
         \label{dist}
     \end{subfigure}
     \begin{subfigure}{0.49\textwidth}
     \centering
         \includegraphics[width=\textwidth]{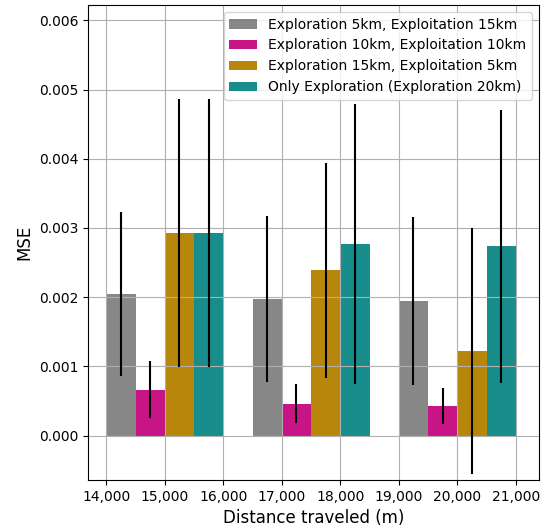}
         \caption{Zoom of the mean of the MSE with 95$\%$ confidence interval of the cases studied for the duration of the exploration and exploitation phases}
         \label{zoomdist}
     \end{subfigure}
     \caption{Results obtained from the cases studied for the duration of the exploration and exploitation phases for four ASVs} 
     \label{explovsexplo}
\end{figure}

Table~\ref{d4v} shows the results of the mean MSE and the confidence interval for four vehicles, obtained with different exploration and exploration distances in the estimation of the water quality model of the entire surface water resource. When the maximum distance that ASVs can travel is 20km, the best combination of exploration and exploration is 10 km and 10 km. This means that the balance of phases to obtain a good model is to explore the surface of the water resource 50$\%$ of the total distance and 50$\%$ to monitor or exploit the contamination zones. Considering the maximum distance equal to 30km, the best result was obtained when the ASVs explored 20km and exploited 10km. However, the difference between the results obtained in the case Exploration 15km and Exploitation 15km case is very low. Therefore, it can be concluded that the best balance between exploration and exploitation is 50$\%$/50$\%$ of the maximum distance, with cases where the distance of the exploration phase can be increased to 67$\%$ of the maximum travel distance. For more detailed results with different fleet sizes, see the \ref{sec:sample:appendix}. The results obtained with four vehicles and a maximum distance of 20km are shown graphically in Fig.~\ref{explovsexplo}.

\begin{table}[H]
\caption{Comparison between the MSE of the distances traveled in the exploration phase and the exploitation phase for four vehicles}
\centering
\resizebox{\textwidth}{!}{
\begin{tabular}{lcccccc}
\hline
Phase distance	& MSE 5km & MSE 10km & MSE 15km & MSE 20km& MSE 25km & MSE 30km\\
\hline
Exploration 5km    	& 0.01388 $\pm$& 0.0265 $\pm$& 0.00217 $\pm$& 0.00209 $\pm$ & 0.00212 $\pm$ & 0.00195 $\pm$\\
Exploitation 25km  &  0.02379 &  0.00702 & 0.00462  &  0.00444 & 0.00447 & 0.00459 \\\hline
\textbf{Exploration 10km }   	& 0.01388 $\pm$ & 0.00427 $\pm$& 0.00068 $\pm$ & \textbf{0.00045 $\pm$}& 0.00045 $\pm$ & 0.00046 $\pm$\\
Exploitation 20km  & 0.02379 & 0.00509  &0.00081  &  \textbf{0.00054} & 0.00064 & 0.00066  \\\hline
\textbf{Exploration 15km }   	& 0.01388 $\pm$ & 0.00427 $\pm$& 0.00337 $\pm$ & 0.00167 $\pm$ & 0.00032 $\pm$ & \textbf{0.00029 $\pm$}	\\
Exploitation 15km  &  0.02379&  0.00509 & 0.00467 & 0.00446 & 0.00017 & \textbf{0.00027 }\\\hline
\textbf{Exploration 20km }  	& 0.01388 $\pm$ & 0.00427 $\pm$& 0.00337 $\pm$ & 0.00299 $\pm$	& 0.00061 $\pm$ & \textbf{0.00025 $\pm$}\\
Exploitation 10km &  0.02379 &  0.00509  &  0.00467 &  0.00418 & 0.00118 & \textbf{0.00033}	\\\hline
Exploration 25km    	&0.01388 $\pm$ & 0.00427 $\pm$& 0.00337 $\pm$ & 0.00299 $\pm$	& 0.00169 $\pm$ & 0.00045 $\pm$		\\
Exploitation 5km &0.02379 &  0.00509  &  0.00467 &  0.00418 & 0.00138 & 0.00137 \\\hline
Exploration 30km    	&  0.01388 $\pm$ & 0.00427 $\pm$& 0.00337 $\pm$ & 0.00299 $\pm$	& 0.00169 $\pm$ & 0.00119 $\pm$		\\
 &0.02379 &  0.00509  &  0.00467 &  0.00418 & 0.00138 & 0.00119 \\\hline
\end{tabular}}
\label{d4v}
\end{table}

For comparison with other path planners, the maximum distance that ASVs can travel is 20km, so the balance between exploration and exploitation that needs to be applied in the monitoring system proposed is 50$\%$/50$\%$. In other words, ASVs explore the surface water resource for 10km and then exploit the access zones for 10km. 

\begin{figure}[H]
    \centering
    \includegraphics[width=\textwidth]{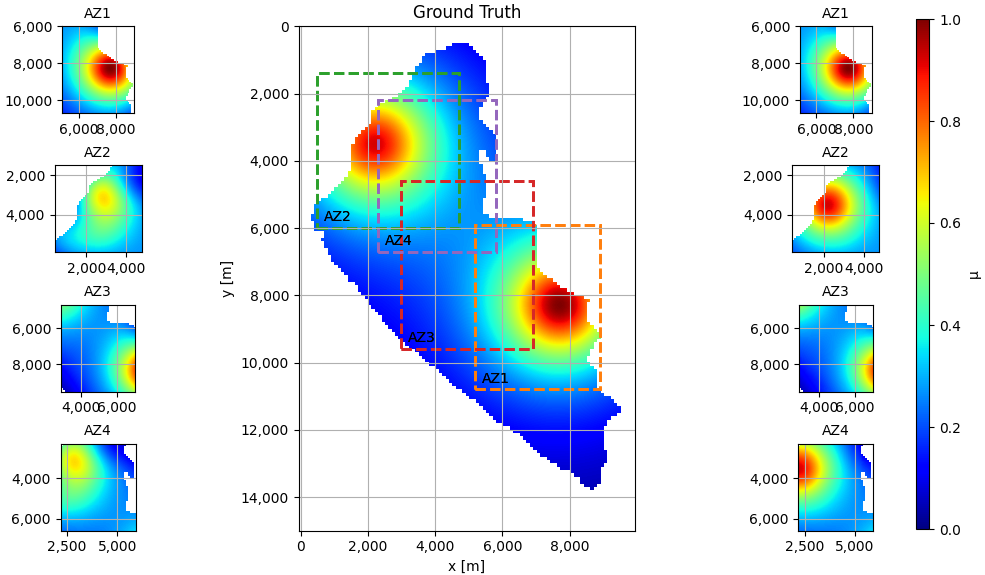}
    \caption{Example of the result obtained with the AquaFeL-PSO monitoring system. The figures on the left show the estimated models of the action zones in the exploration phase, the figure in the center represents the ground truth of the Ypacarai lake, and the figures on the right represent the estimated models of the action zones after the exploitation phase (final model of the action zones).}
    \label{zoommulti}
\end{figure}

Fig.~\ref{zoommulti} shows the comparison between the model obtained from the action zone in the exploration phase (left), the ground truth (center), and the model obtained from the action zones in the exploitation phase (right). For the exploration phase, the vehicles traveled 10km over the surface of the lake. Likewise, in the exploitation phase, the ASVs exploited the contamination zones until they reached a traveled distance of 10km. This figure shows the improvement of the estimated model in the contamination zones, mainly in the action zone 1 and action zone 2.

\subsubsection{Centralized vs Federated Learning}

This subsection shows the comparison between the application of the FL technique and a centralized learning. Using the centralized learning, the samples taken by the fleet in the exploitation phase go directly to the main server, and a single estimated water quality model is generated, unlike the FL. The results show in Table~\ref{tab:cenvfed} demonstrate that there is very little difference between using a centralized learning or applying the FL technique. However, as an advantage, with FL, there is no excess data on the central server, since the GP is adjusted at the nodes in each action zone.

\begin{table}[H]
  \caption{Comparison of MSE and error applying centralized learning and federated learning techniques}
  \label{tab:cenvfed}
  \resizebox{\textwidth}{!}{ 
  \centering
  \begin{tabular}{lcccc}
    \hline
    Technique\ & \ MSE (Action\ &\ Error (Peaks)\  &\ MSE (Model of the & Mean of Time\\
    & Zones)\ &\  & Lake) & Consuming (sec)\\
    \hline
    Centralized Learning & 0.02304 $\pm$ 0.05757  & 0.00155 $\pm$ 0.00985 & 0.00038 $\pm$ 0.00063 & 8.43176 $\pm$ 2.67029\\ \hline
    Federated Learning & 0.02331 $\pm$ 0.05749 & 0.00160 $\pm$ 0.00919 & 0.00045 $\pm$ 0.00055 & 8.29892 $\pm$ 2.22732 \\\hline
\end{tabular}}
\end{table}

\subsection{Performance comparison}

After making the necessary adjustments, the performance of the path planners is compared. It is used as a basis algorithm developed by \cite{sakai2018pythonrobotics} for the Lawn mower path planner. The settings for the compared path planners, with respect to water sampling and GP, are the same settings as shown in Sect.~\ref{settings}. The maximum distance traveled by the vehicles is 20,000 meters. To set the values of the hyper-parameters in the Classic PSO, \cite{xin2019application}\cite{cui2020multi} are taken into consideration, $c_1 = c_2 = 2$. For the Enhanced GP-based PSO (Exploration), the values of the coefficients are the same as those used in the exploration phase, Table~\ref{coefphases}. Moreover, the values of the coefficients used in the Enhanced GP-based PSO (Exploitation) have the same values used in the exploitation phase, Table~\ref{coefphases}. The best combination of parameters for the epsilon greedy method is shown in the Table~\ref{tab:valuesEP}.  Finally, the values of the coefficients for the AquaFeL-PSO algorithm are those shown in the Table~\ref{coefphases}.

\begin{table}[H]
    \caption{Parameter and hyper-parameter values of the Epsilon Greedy method}
    \centering
    \begin{tabular}{cc}
        \hline
        Parameter/Hyper-parameter	&Value \\
        \hline
        $d\epsilon_0$ (m)   	& 6,500 	\\ \hline
        $d\epsilon_f$ (m)  	& 13,500 	\\ \hline
        $\Delta\epsilon$   	&  0.13 	\\\hline
        $c_{1explore}$ & 2.0187 	\\\hline
        $c_{2explore}$ & 0 	\\\hline
        $c_{3explore}$ & 3.2697 	\\\hline
        $c_{4explore}$ & 0 	\\ \hline
        $c_{1exploit}$ & 3.6845 	\\\hline
        $c_{2exploit}$ & 1.5614 	\\\hline
        $c_{3exploit}$ &  0 	\\\hline
        $c_{4exploit}$ & 3.1262 	\\
        \hline
    \end{tabular}
    \label{tab:valuesEP}
\end{table}

The comparison of the mean MSE and errors together with the 95$\%$ confidence interval is shown in Table~\ref{tab:mseall}. The results of the table show that: 1) in the three cases considered (MSE of the action zone, error of the peaks, and MSE of the whole lake), the proposed monitoring system has the best performance; 2) the difference between the models obtained from the action zones of the AquaFeL-PSO algorithm and the other path planners is not large, however, in the detection of the peak of each action zone there are large differences, with the epsilon greedy being the second lowest error, the error obtained with this path planner is more than 40 times greater than the error obtained with the AquaFeL-PSO algorithm; and 3) with respect to the estimated model of the whole lake, the best MSE has the AquaFeL-PSO algorithm and the second best is the MSE of the Enhanced GP-based PSO based on the epsilon greedy method, the difference between the proposed monitoring system and the epsilon greedy method is more than 400$\%$, and the result obtained with the lawn mower algorithm is approximately 70 times greater than the AquaFeL-PSO algorithm.

\begin{table}[H]
    \caption{Comparison of the MSE and the error of the path planners for four vehicles}
    \label{tab:mseall}
    \resizebox{\textwidth}{!}
    { 
        \centering
        \begin{tabular}{lccc}
            \hline
            Path Planner\ & \ MSE (Action\ &\ Error (Peaks)\  &\ MSE (Model of the\\
            & Zones)\ &\  & Lake) \\
            \hline
            Lawn mower & 0.07988 $\pm$ 0.30728 & 0.22435 $\pm$ 0.64639 & 0.03225 $\pm$ 0.07159 \\\hline
            Classic PSO & 0.04095 $\pm$ 0.09118 & 0.14414 $\pm$ 0.57829 & 0.01146 $\pm$ 0.02812\\ \hline
            Enhanced GP-based PSO & 0.02642 $\pm$ 0.05389 & 0.10634 $\pm$ 0.32627 & 0.00268 $\pm$ 0.00388\\ 
            (Exploration) &&&\\\hline
            Enhanced GP-based PSO & 0.04403 $\pm$ 0.10310 & 0.15388 $\pm$ 0.59155 & 0.01266 $\pm$ 0.03218\\ 
            (Exploitation) &&&\\\hline
            Epsilon Greedy Method & 0.02648 $\pm$ 0.05869 & 0.07599 $\pm$ 0.31318 & 0.00246 $\pm$ 0.00760 \\\hline
            \textbf{AquaFeL-PSO} & \textbf{0.02331 $\pm$ 0.05749} & \textbf{0.00160 $\pm$ 0.00919} & \textbf{0.00045 $\pm$ 0.00055} \\\hline
        \end{tabular}
    }
\end{table}

In Fig~\ref{fig:comparison}, the movement of a four-vehicle fleet and the estimated water quality models of the Ypacarai lake can be seen applying different path planners. The graphs at the top show the trajectory performed by the ASVs and the uncertainty of the model. The trajectory of each ASV is represented by a different color. In addition, the initial and final positions of each ASV are indicated by black and red dots, respectively. The graphs at the bottom show the estimated water quality model of the Ypacarai lake. The actual model of this scenario is the ground truth shown in Fig.~\ref{ground}.

Fig.~\ref{lm} shows how the ASVs move across the surface of the water resource by applying the Lawn mower path planner. When there are no limited factors, such as time and distance, this path planner is able to cover the entire surface of the lake. However, when considering a limited distance, the ASVs are not able to obtain enough samples to generate an accurate model, nor are they able to detect pollution peaks. This leads to large errors in the estimation of the water quality model. Fig.~\ref{subfig:cla} represents the trajectories generated by the classical PSO. In the Classic PSO, the ASVs are guided by their own experience and by the experience of the other vehicles in the fleet, because of this, after some time, the vehicles are stuck in a local optimum, since they cannot explore the surface anymore. In this scenario, the influence of the global best on the vehicles at the bottom of the lake can be observed, where both vehicles are in contaminated areas. However, as another of the ASVs has found the maximum peak, the global best is the best position of that vehicle, and the ASVs that were at the bottom of the lake, instead of exploiting the area where they are positioned, go to the area where the maximum peak is located. In Fig.~\ref{egper}, it is shown how ASVs explore the zones where uncertainty is high and are not guided by the zones where contamination is high. This is due to the values of the acceleration coefficients, and the terms that are active are the local best and the maximum uncertainty term. Due to the local best, the vehicles exploit the areas through which they pass, and thanks to the maximum uncertainty, the ASVs go to unexplored areas. This behavior allows to obtain a good water quality model of the water resource. On the other hand, it does not exploit the contamination zones, and as a consequence, a considerable error in the contamination peaks can be obtained. In contrast to Fig.~\ref{egper}, in Fig.~\ref{egpet}, the movement of vehicles is shown by applying the Enhanced GP-based PSO focused on the exploitation of pollution zones. The active terms in this path planner are the local best, the global best and maximum contamination. Since the ASVs do not explore the surface of the water resource sufficiently, the vehicles are not able to detect all pollution peaks, so the maximum contamination value of the $\bb{max\_con}$ term will be determined by the samples they have taken so far. This means that, when one of the vehicles finds an area with high values, the ASVs will go to that point, whether this is the area where the maximum contamination peak is found or not. Fig~\ref{egm} shows the trajectory of the ASVs obtained by applying the epsilon greedy method, where, during monitoring, the algorithm changes the focus randomly. At the beginning of the monitoring, the probability that the algorithm explores the surface is high. However, as time passes, the probability of exploring the surface decreases and the probability of exploiting contamination zones increases. That can be observed in Fig.~\ref{egm}, as the ASVs target the area where the generated model estimates the maximum contamination point. As the change of approach is random, there are cases where not enough data are taken to generate a good estimated model, which can affect the detection of contamination zones. Fig.~\ref{multipso} shows the movements of the fleet applying the proposed monitoring system. Since the system considers two phases in given periods, and the search space is divided into action zones as well as the fleet into subpopulations, it is possible to obtain an accurate model of the pollution zones and the whole lake. Moreover, the path planner is able to detect pollution peaks. After performing the explore of the surface of the lake, the model generated with the water samples is obtained, the model is as shown in Fig.~\ref{4v}. Then, with this model, the algorithm divides the map into action zones, as shown in Fig.~\ref{4vl}. Once the action zones are calculated, the vehicles are assigned. In this scenario, each ASV is assigned an action zone. Finally, the vehicles are dedicated to exploit the contamination zones obtaining the models of each action zone. Once the exploitation phase is completed, the algorithm proceeds to replace the results obtained from the action zones with the estimated model generated in the exploration phase, creating the final water quality model of the water resource shown in Fig.~\ref{multipso}.

\begin{figure}
\centering
     \begin{subfigure}{0.35\textwidth}
     \centering
         \includegraphics[width=0.9\textwidth]{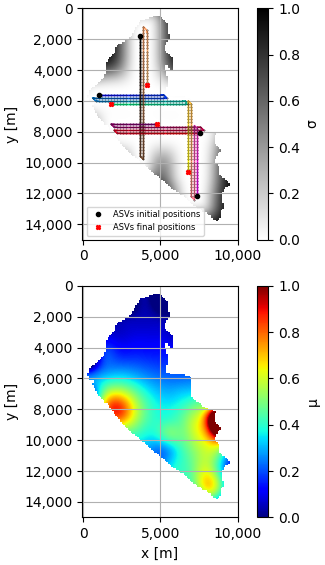}
         \caption{Lawn mower}
         \label{lm}
     \end{subfigure}
     \begin{subfigure}{0.35\textwidth}
     \centering
         \includegraphics[width=0.9\textwidth]{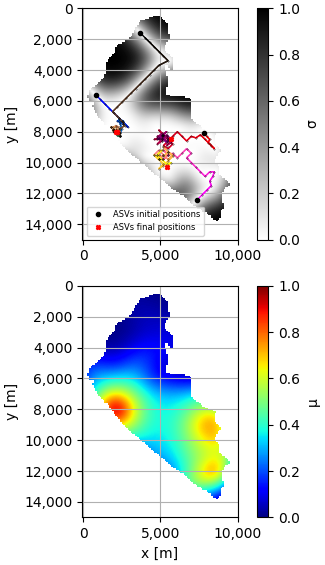}
         \caption{Classic PSO}
         \label{subfig:cla}
     \end{subfigure}
     \begin{subfigure}{0.35\textwidth}
     \centering
         \includegraphics[width=0.9\textwidth]{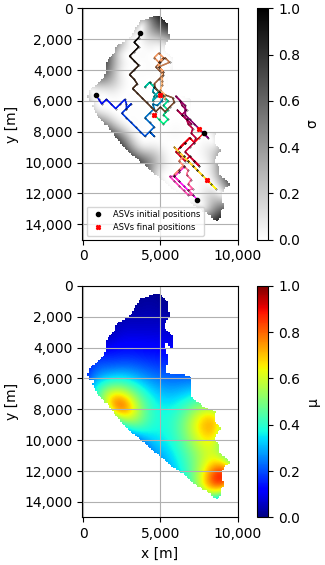}
         \caption{Enhanced GP-based PSO (Exploration)}
         \label{egper}
     \end{subfigure}
     \begin{subfigure}{0.35\textwidth}
     \centering
         \includegraphics[width=0.9\textwidth]{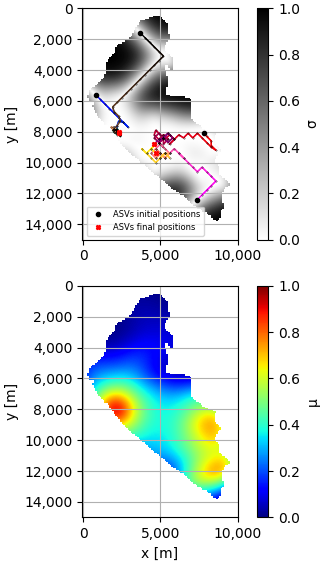}
         \caption{Enhanced GP-based PSO (Exploitation)}
         \label{egpet}
     \end{subfigure}
\end{figure}

\begin{figure}
\ContinuedFloat
    \centering
         \begin{subfigure}{0.35\textwidth}
     \centering
         \includegraphics[width=0.9\textwidth]{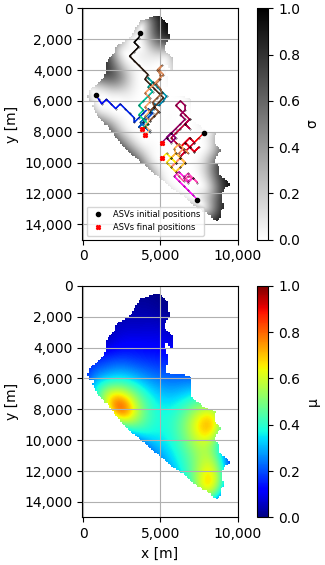}
         \caption{Epsilon Greedy method}
         \label{egm}
     \end{subfigure}
     \begin{subfigure}{0.35\textwidth}
     \centering
         \includegraphics[width=0.9\textwidth]{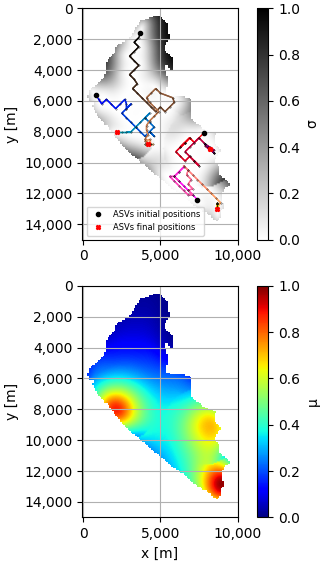}
         \caption{AquaFeL-PSO}
         \label{multipso}
     \end{subfigure}
    \caption{Representation of the results obtained in the monitoring of the Ypacarai lake. The graphs at the top represent the movement of the ASVs and the model uncertainty, and the graphs at the bottom represent the estimated water quality model of the lake.} 
    \label{fig:comparison}
    
\end{figure}

\section{Discussion of the Results}\label{sec:disc}

The main findings of this work are discussed below:

\begin{itemize}
    \item The scalability and vehicle allocation results showed that the proposed monitoring system works well with various fleet sizes. Moreover, it was verified that the vehicle assignment works well according to the priority of the action zone and the distance between the action center and the vehicle.
    \item Regarding the duration of exploration and exploitation phases, the tests showed that the proposed system performed well when the balance between the duration of the phases is between 50$\%$/50$\%$ and 67$\%$/37$\%$ of the maximum travel distance.
    \item Comparisons were made between a centralized learning-based system and the proposed FL-based system, and the results of model estimation and contamination peak detection for both systems turned out to be quite similar. However, the FL allows the reduction of data traffic between local nodes and the main server. In addition, in isolated areas, there is a possibility of having connection problems with the main server. Whereas, by applying the FL technique, the GP tuning will be done by local nodes.
    \item The lawn mower algorithm, having a maximum distance that the ASVs can travel, did not manage to explore the entire surface of the lake and, as a consequence, was not able to detect pollution peaks or to generate good models of the action zones and of the entire water resource.
    \item As water resource monitoring is a multimodal problem, the Classic PSO was not able to detect all pollution peaks, since it is stuck in a local optimum. This made the estimated water quality model unreliable.
    \item The Enhanced GP-based PSO focused on exploration generated a good estimated model of the entire lake. However, it was not able to detect the peaks of water resource contamination.
    \item The Enhanced GP-based PSO focused on exploitation and was capable of detecting the maximum contamination peak. However, having several local and global peaks, the algorithm was not able to detect all peaks. Moreover, as the focus is exploitation, it did not generate good estimated models of water quality neither for the pollution zones, nor for the whole lake.
    \item The Epsilon greedy method performed well with respect to the three cases studied. This is due to the change of approach during the monitoring task. The algorithm was able to detect the contamination peaks and generate good estimated models.
    \item The proposed monitoring system, the AquaFeL-PSO, performed the best among all path planners compared. The water quality model of the contaminated areas obtained by the AquaFeL-PSO is 14$\%$ better than the Enhanced GP-based PSO, the second best MSE. In addition, the water quality model of the whole water resource obtained by the AquaFel-PSO is at least 400$\%$ better than the second best MSE, the Epsilon Greedy method. Regarding the detection of contamination peaks, the results obtained with the proposed monitoring system have an improvement of more than 4000$\%$. This is because the algorithm: has two controlled phases; and it divides the map into specific zones for vehicles to exploit, causing the fleet to be divided into subpopulations. 
\end{itemize}

\section{Conclusions}\label{sec:conclusion}

A water resource monitoring system for ASVs was developed and simulated in this paper. The designed monitoring system is based on a multimodal PSO and federated learning technique and uses the Gaussian process as a surrogate model. The objectives of the proposed monitoring system were to generate accurate estimated models of the contamination zones and the whole surface of the water resource, as well as the detection of contamination peaks. The AquaFeL-PSO algorithm has two phases, the exploration phase and the exploitation phase. In the exploration phase, the ASVs are dedicated to cover the surface of the water resource to obtain water samples from the water body and generate a first estimated water quality model. In the exploitation phase, using this first model generated, the surface of the water body is divided into action zones according to the length of the water resource and the number of vehicles in the fleet. Then, the ASVs are assigned to the action zones to carry out the exploitation of these areas. For the estimation of the water quality model of the action zones, the FL technique is used, which allows alleviating the data traffic between the nodes and the central server. Each action zone has its own node, and in each node the estimated model of that area is updated. Finally, when the monitoring task is completed, the models of the action zones are joined with the model obtained in the exploration phase to obtain the final water quality model of the water resource. It was demonstrated that with the AquaFeL-PSO it is possible to solve the multimodal problem of water resource monitoring. The proposed monitoring system was able to detect pollution peaks and obtain accurate estimated water quality models of the pollution zones and the entire surface of the water body. As future work, the monitoring system developed will be improved to be able to determine the speed and direction of movement of the contamination zones in dynamic environments. Furthermore, a multiobjective PSO will be employed to obtain multiple water quality parameter models simultaneously. Therefore, the federated learning approach will be extended to the estimation of multiple models.

Funding: This work has been partially funded by the regional government Junta de Andalucía under the Projects “Despliegue Inteligente de una red de Vehículos Acuáticos no Tripulados para la Monitorización de Recursos Hídricos US-1257508' and 'Despliegue y Control de una red Inteligente of Vehculos Autónomos Acuáticos for the monitorisation of Recursos Hdricos Andaluces PY18-RE0009'.

\appendix

\section{Appendix}\label{sec:sample:appendix}

This section shows the results obtained with two, six, and eight vehicles. 

\subsection{Proposed approach evaluation}

\subsubsection{Exploration vs Exploitation duration}

This subsection details the results obtained for two (Table~\ref{d2v}), six (Table~\ref{d6v}) and eight vehicles (Table~\ref{d8v}) with respect to the duration of the exploration phase and the exploitation phase. Table~\ref{d2v} shows the results obtained with two ASVs, when the maximum distance that ASVs can travel is 20km, the best combination of exploration and exploration is 10 km and 10 km. This same result is obtained for six (Table~\ref{d6v}) and eight ASVs (Table~\ref{d8v}). Considering the maximum distance equal to 30km, the best result with two vehicles (Table~\ref{d2v}) was obtained in the Exploration 25km and Exploitation 5km. However, by a small margin, the second best was Exploration 15km and Exploitation 15km. For eight ASVs ((Table~\ref{d8v})), similar results were obtained with Exploration 10km and Exploitation 20km, Exploration 15km and Exploitation 15km, and Exploration 20km and Exploitation 10km. It can be said that, with the balance between exploration and exploitation equal to 50$\%$/50$\%$, a good performance of the algorithm is obtained. The best result with six ASVs (Table~\ref{d6v}) was obtained when the ASVs explored 20km and exploited 10km. However, the difference between the results obtained in the case Exploration 15km and Exploitation 15km case is very low. Therefore, it can be concluded that the best balance between exploration and exploitation is 50$\%$/50$\%$ of the maximum distance, with cases where the distance of the exploration phase can be increased to 67$\%$ of the maximum travel distance. From the results obtained in Table~\ref{d8v}, it can be concluded that, having more vehicles, the difference is minimal whether the ASVs travel 20km or 30km. Therefore, the ideal distance for eight ASVs is 20km, since there is no significant difference in the results.

\begin{table}[H]
\caption{Comparison between the MSE of the distances traveled in the exploration phase and the exploitation phase for two vehicles}
\centering
\resizebox{0.98\textwidth}{!}{
\begin{tabular}{lcccccc}
\hline
Vs	& MSE 5km & MSE 10km & MSE 15km & MSE 20km & MSE 25km & MSE 30km\\
\hline
Exploration 5km    	& 0.06180 $\pm$ & 0.04020 $\pm$ & 0.03476 $\pm$ & 0.03425 $\pm$ &  0.03403 $\pm$ & 0.03395 $\pm$		\\
Exploitation 25km  & 0.05052 & 0.03669 & 0.03538 & 0.03635 & 0.03646 & 0.03656	\\\hline
\textbf{Exploration 10km }   	& 0.06180 $\pm$ & 0.02646 $\pm$ & 0.01610 $\pm$ & \textbf{0.00622 $\pm$} & 0.00574 $\pm$ & 0.00562 $\pm$	\\
Exploitation 20km  & 0.05052 & 0.03128 & 0.02942 & \textbf{0.02236} & 0.01538 & 0.01123  	 \\\hline
\textbf{Exploration 15km}   	& 0.06180 $\pm$ & 0.02646 $\pm$ & 0.01762 $\pm$ & 0.01492 $\pm$	& 0.00679 $\pm$ & \textbf{0.00375 $\pm$}	\\
Exploitation 15km  & 0.05052 & 0.03128 & 0.02716 & 0.02734 & 0.01484 & \textbf{0.01047} 	 \\\hline
Exploration 20km    	& 0.06180 $\pm$ & 0.02646 $\pm$ & 0.01762 $\pm$ & 0.01634 $\pm$	& 0.01231 $\pm$ & 0.00611 $\pm$	\\
Exploitation 10km & 0.05052 & 0.03128 & 0.02716 & 0.02800& 0.02511 & 0.01835 	\\\hline
\textbf{Exploration 25km  }  	&0.06180 $\pm$ & 0.02646 $\pm$ & 0.01762 $\pm$ & 0.01634 $\pm$	& 0.01253 $\pm$ & \textbf{0.00338 $\pm$	}	\\
Exploitation 5km & 0.05052 & 0.03128 & 0.02716 & 0.02800 & 0.02214 & \textbf{0.00857} \\\hline
Exploration 30km    	&  0.06180 $\pm$ & 0.02646 $\pm$ & 0.01762 $\pm$ & 0.01634 $\pm$	& 0.01253 $\pm$ & 0.01077 $\pm$		\\
 &0.05052 & 0.03128 & 0.02716 & 0.02800 & 0.02214 & 0.01879 \\\hline
\end{tabular}}
\label{d2v}
\end{table}

\begin{table}[H]
\caption{Comparison between the MSE of the distances traveled in the exploration phase and the exploitation phase for six vehicles}
\centering
\resizebox{\textwidth}{!}{
\begin{tabular}{lcccccc}
\hline
Phase distance	& MSE 5km & MSE 10km & MSE 15km & MSE 20km& MSE 25km & MSE 30km\\
\hline
Exploration 25km    	& 0.00704 $\pm$ & 0.00142 $\pm$& 0.00095 $\pm$ & 0.00094 $\pm$ &	0.00093 $\pm$ &	0.00093 $\pm$\\
Exploitation 15km  &0.00701  & 0.00329  &  0.00198 & 0.00197 & 0.00198 & 0.00198 \\\hline
\textbf{Exploration 10km}    	& 0.00704 $\pm$& 0.00276 $\pm$ & 0.00065 $\pm$ & \textbf{0.00042 $\pm$} &	0.00040 $\pm$ &	0.00040 $\pm$ \\
Exploitation 20km  & 0.00701  & 0.00628 &0.00194  & \textbf{0.00097} & 0.00092 & 0.00094 \\\hline
\textbf{Exploration 15km }   	& 0.00704 $\pm$ & 0.00276 $\pm$ & 0.00201 $\pm$ & 0.00060 $\pm$&	0.00050 $\pm$ &	\textbf{0.00039 $\pm$}	\\
Exploitation 15km  & 0.00701 & 0.00628 & 0.00538 &  0.00162& 0.00168 & \textbf{0.00128} \\\hline
\textbf{Exploration 20km}   	& 0.00704 $\pm$ & 0.00276 $\pm$& 0.00201 $\pm$ & 0.00136 $\pm$& 0.00043	 $\pm$ &	\textbf{0.00033 $\pm$}	\\
Exploitation 10km &  0.00701 &  0.00628  & 0.00538  &  0.00328  & 0.00103 & \textbf{0.00085 }	\\\hline
Exploration 25km    	&0.00704 $\pm$ & 0.00276 $\pm$& 0.00201 $\pm$ & 0.00136 $\pm$& 0.00113 $\pm$ & 0.00052 $\pm$		\\
Exploitation 5km & 0.00701 &  0.00628  & 0.00538  &  0.00328 & 0.00302 & 0.00149 \\\hline
Exploration 30km    	&   0.00704 $\pm$ & 0.00276 $\pm$& 0.00201 $\pm$ & 0.00136 $\pm$& 0.00113 $\pm$ & 0.00092 $\pm$		\\
 &0.00701 &  0.00628  & 0.00538  &  0.00328 & 0.00302 & 0.00299 \\\hline
\end{tabular}}
\label{d6v}
\end{table}

\begin{table}[H]
\caption{Comparison between the MSE of the distances traveled in the exploration phase and the exploitation phase for eight vehicles}
\centering
\resizebox{0.98\textwidth}{!}{
\begin{tabular}{lcccccc}
\hline
Phase distance	& MSE 5km & MSE 10km & MSE 15km & MSE 20km& MSE 25km & MSE 30km\\
\hline
Exploration 5km    	& 0.00137 $\pm$ & 0.00041 $\pm$ & 0.00029 $\pm$ & 0.00028 $\pm$&	0.00028 $\pm$ &	0.00028 $\pm$	\\
Exploitation 25km  & 0.00196 & 0.00090 & 0.00056 &  0.00054& 0.00054 & 0.00054 \\\hline
\textbf{Exploration 10km }   	& 0.00137 $\pm$ & 0.00023 $\pm$& 0.00006 $\pm$  & \textbf{0.00003 $\pm$ }&	0.00002 $\pm$ &\textbf{	0.00002 $\pm$ }\\
Exploitation 20km  &  0.00196&  0.00044 & 0.00013 &\textbf{0.00003 }& 0.00003 & \textbf{0.00003}\\\hline
\textbf{Exploration 15km  }  	& 0.00137 $\pm$  & 0.00023 $\pm$& 0.00011 $\pm$ & \textbf{0.00003 $\pm$}& 0.00002	 $\pm$ & \textbf{0.00002	 $\pm$}		\\
Exploitation 15km  & 0.00196&  0.00044 & 0.00035 &\textbf{ 0.00004 }& 0.00003 & \textbf{0.00003 }\\\hline
\textbf{Exploration 20km  }  	& 0.00137 $\pm$ & 0.00023 $\pm$& 0.00011 $\pm$  & 0.00006 $\pm$& 0.00005	 $\pm$ &	\textbf{0.00002 $\pm$}		\\
Exploitation 10km &  0.00196 &  0.00044  & 0.00035 & 0.00018 & 0.00019  & \textbf{0.00004  }	\\\hline
Exploration 25km    	&0.00137 $\pm$ & 0.00023 $\pm$& 0.00011 $\pm$  & 0.00006 $\pm$& 0.00005	 $\pm$ & 0.00006 $\pm$		\\
Exploitation 5km &0.00196 &  0.00044  & 0.00035 & 0.00018 & 0.00018 & 0.00028 \\\hline
Exploration 30km    	& 0.00137 $\pm$ & 0.00023 $\pm$& 0.00011 $\pm$  & 0.00006 $\pm$& 0.00005	 $\pm$ & 0.00005 $\pm$		\\
 &0.00196 &  0.00044  & 0.00035 & 0.00018 & 0.00018 & 0.00019 \\\hline
\end{tabular}}
\label{d8v}
\end{table}

\subsection{Performance comparison}

The Tables~\ref{tab:mseall2} and \ref{tab:mseall6} show the results obtained from the comparison of the path planners. Having a fleet with two vehicles, in the three case studies, the path planner that obtained the best performance was the proposed path planner. Unlike the results obtained with four vehicles, the difference in errors is not so large. However, the MSE of the estimated model of the whole lake and the error in the pollution peaks are up to 4 times lower than the second best MSE and error. With respect to the results obtained with six ASVs, in the modeling of the action zones, with a very small difference, the proposed path planner obtained the second best performance, behind the epsilon greedy method. However, in the other two case studies, the AquaFeL-PSO algorithm performed the best. In the detection of pollution peaks, the second best result is almost 20 times higher than the error obtained with the AquaFeL-PSO algorithm, and in the generation of the estimated model of the whole lake, the second best MSE is about 200$\%$ higher than the MSE of the proposed monitoring system.

\begin{table}[H]
  \caption{Comparison of the MSE and the error of the path planners for two vehicles}
  \label{tab:mseall2}
  \resizebox{\textwidth}{!}{ 
  \centering
  \begin{tabular}{lccc}
    \hline
    Path Planner\ & \ MSE (Action\ &\ Error (Peaks)\  &\ MSE (Model of the\\
    & Zones)\ &\  & Lake) \\
    \hline
    Lawn mower & 0.06407 $\pm$ 0.11914 & 0.56516 $\pm$ 0.57607 & 0.04625 $\pm$ 0.03788 \\\hline
    Classic PSO & 0.05912 $\pm$ 0.11250 & 0.33411 $\pm$ 0.75806 & 0.04272 $\pm$ 0.05509 \\\hline
    Enhanced GP-based PSO & 0.03577 $\pm$ 0.06286 & 0.33622 $\pm$ 0.76587 & 0.01631 $\pm$ 0.02798 \\
    (Exploration) &&&\\\hline
    Enhanced GP-based PSO & 0.06164 $\pm$ 0.11393 & 0.36485 $\pm$ 0.80912 & 0.04378 $\pm$ 0.05145 \\
    (Exploitation) &&&\\\hline
    Epsilon Greedy Method & 0.04524 $\pm$ 0.08610 & 0.39966 $\pm$ 0.85490 & 0.03297 $\pm$ 0.04059 \\\hline
    AquaFel-PSO & \textbf{0.02361 $\pm$ 0.05634} & \textbf{0.07770 $\pm$ 0.44700} & \textbf{0.00526 $\pm$ 0.011230} \\\hline
\end{tabular}}
\end{table}

\begin{table}[H]
  \caption{Comparison of the MSE and the error of the path planners for six vehicles}
  \label{tab:mseall6}
  \resizebox{\textwidth}{!}{ 
  \centering
  \begin{tabular}{lccc}
    \hline
    Path Planner\ & \ MSE (Action\ &\ Error (Peaks)\  &\ MSE (Model of the\\
    & Zones)\ &\  & Lake) \\
    \hline
    Lawn mower & 0.21101 $\pm$ 1.61966 & 0.27808 $\pm$ 1.20208 & 0.05738 $\pm$ 0.05048 \\\hline
    Classic PSO & 0.03357 $\pm$ 0.07676 & 0.06726 $\pm$ 0.25293 & 0.00450 $\pm$ 0.00609 \\\hline
    Enhanced GP-based PSO & 0.03472 $\pm$ 0.07295 & 0.04969 $\pm$ 0.19958 & 0.00131 $\pm$ 0.00313\\ 
    (Exploration) &&&\\\hline
    Enhanced GP-based PSO & 0.03667 $\pm$ 0.08129 & 0.06685 $\pm$ 0.25677 & 0.00501 $\pm$ 0.00770\\ 
    (Exploitation) &&&\\\hline
    Epsilon Greedy Method & \textbf{0.03122 $\pm$ 0.06354} & 0.05911 $\pm$ 0.25621 & 0.00200 $\pm$ 0.00385\\\hline
    AquaFel-PSO & \textbf{0.03156 $\pm$ 0.06709} & \textbf{0.00255 $\pm$ 0.01590} & \textbf{0.00042 $\pm$ 0.00097} \\\hline
\end{tabular}}
\end{table}

\bibliographystyle{elsarticle-num} 
\bibliography{reference}

\newcommand{\noop}[1]{}
\begin{thebibliography}{10}
\expandafter\ifx\csname url\endcsname\relax
  \def\url#1{\texttt{#1}}\fi
\expandafter\ifx\csname urlprefix\endcsname\relax\def\urlprefix{URL }\fi
\expandafter\ifx\csname href\endcsname\relax
  \def\href#1#2{#2} \def\path#1{#1}\fi

\bibitem{desa2016transforming}
U.~Desa, et~al., Transforming our world: The 2030 agenda for sustainable
  development, United Nations (2016).

\bibitem{itaipu2018}
{Dirección General del Centro Multidisciplinario de Investigaciones
  Tecnológicas (CEMIT)}, Servicios de monitoreo de calidad de agua por
  campañas de muestreo en el lago ypacaraí. 2016 -2018, Tech. rep.,
  Universidad Nacional de Asunción (UNA) (2018).

\bibitem{itaipu2021}
{Dirección General del Centro Multidisciplinario de Investigaciones
  Tecnológicas (CEMIT)}, Monitoreo de calidad de agua por campañas de
  muestreo en el lago ypacaraí 2019 - 2021, Tech. rep., Universidad Nacional
  de Asunción (UNA) (2021).

\bibitem{lopez2018eutrophication}
G.~A. L{\'o}pez~Moreira, L.~Hinegk, A.~Salvadore, G.~Zolezzi, F.~H{\"o}lker,
  R.~A. Monte Domecq~S, M.~Bocci, S.~Carrer, L.~De~Nat, J.~Escrib{\'a}, et~al.,
  Eutrophication, research and management history of the shallow ypacara{\'\i}
  lake (paraguay), Sustainability 10~(7) (2018) 2426.

\bibitem{arzamendia2019intelligent}
M.~Arzamendia, D.~G.~Reina, S.~Toral, D.~Gregor, E.~Asimakopoulou, N.~Bessis,
  Intelligent online learning strategy for an autonomous surface vehicle in
  lake environments using evolutionary computation, IEEE Intelligent
  Transportation Systems Magazine 11~(4) (2019) 110--125.

\bibitem{luis2021multiagent}
S.~Y. {Luis}, D.~G. {Reina}, S.~L.~T. {Marín}, A multiagent deep reinforcement
  learning approach for path planning in autonomous surface vehicles: The
  ypacarac-lake patrolling case., IEEE Access (2021).

\bibitem{lin2022smart}
C.~Lin, G.~Han, T.~Zhang, S.~B.~H. Shah, Y.~Peng, Smart underwater pollution
  detection based on graph-based multi-agent reinforcement learning towards
  auv-based network its, IEEE Transactions on Intelligent Transportation
  Systems (2022).

\bibitem{peralta2021}
F.~Peralta, D.~G. Reina, S.~Toral, M.~Arzamendia, D.~Gregor, A bayesian
  optimization approach for water resources monitoring through an autonomous
  surface vehicle: The ypacarai lake case study, IEEE Access 9 (2021)
  9163--9179.
\newblock \href {https://doi.org/10.1109/ACCESS.2021.3050934}
  {\path{doi:10.1109/ACCESS.2021.3050934}}.

\bibitem{panetsos2022vision}
F.~Panetsos, P.~Rousseas, G.~Karras, C.~Bechlioulis, K.~J. Kyriakopoulos, A
  vision-based motion control framework for water quality monitoring using an
  unmanned aerial vehicle, Sustainability 14~(11) (2022) 6502.

\bibitem{sanchez2019distributed}
J.~S{\'a}nchez-Garc{\'\i}a, D.~G. Reina, S.~Toral, A distributed pso-based
  exploration algorithm for a uav network assisting a disaster scenario, Future
  Generation Computer Systems 90 (2019) 129--148.

\bibitem{song2017global}
B.~Song, Z.~Wang, L.~Zou, On global smooth path planning for mobile robots
  using a novel multimodal delayed pso algorithm, Cognitive Computation 9~(1)
  (2017) 5--17.

\bibitem{bottarelli2019orienteering}
L.~Bottarelli, M.~Bicego, J.~Blum, A.~Farinelli, Orienteering-based informative
  path planning for environmental monitoring, Engineering Applications of
  Artificial Intelligence 77 (2019) 46--58.

\bibitem{yang2019federated}
Q.~Yang, Y.~Liu, Y.~Cheng, Y.~Kang, T.~Chen, H.~Yu, Federated learning,
  Synthesis Lectures on Artificial Intelligence and Machine Learning 13~(3)
  (2019) 1--207.

\bibitem{chen2020wireless}
M.~Chen, H.~V. Poor, W.~Saad, S.~Cui, Wireless communications for collaborative
  federated learning, IEEE Communications Magazine 58~(12) (2020) 48--54.

\bibitem{ye2020federated}
D.~Ye, R.~Yu, M.~Pan, Z.~Han, Federated learning in vehicular edge computing: A
  selective model aggregation approach, IEEE Access 8 (2020) 23920--23935.

\bibitem{peralta2021bayesian}
F.~Peralta, D.~G. Reina, S.~Toral, M.~Arzamendia, D.~Gregor, A bayesian
  optimization approach for multi-function estimation for environmental
  monitoring using an autonomous surface vehicle: Ypacarai lake case study,
  Electronics 10~(8) (2021) 963.

\bibitem{kathen2021informative}
M.~J. Ten~Kathen, I.~J. Flores, D.~G. Reina, An informative path planner for a
  swarm of asvs based on an enhanced pso with gaussian surrogate model
  components intended for water monitoring applications, Electronics 10~(13)
  (2021) 1605.

\bibitem{luis2020deep}
S.~Y. Luis, D.~G. Reina, S.~L.~T. Mar{\'\i}n, A deep reinforcement learning
  approach for the patrolling problem of water resources through autonomous
  surface vehicles: The ypacarai lake case, IEEE Access 8 (2020)
  204076--204093.

\bibitem{holland1992genetic}
J.~H. Holland, Genetic algorithms, Scientific american 267~(1) (1992) 66--73.

\bibitem{arzamendia2019evolutionary}
M.~Arzamendia, D.~Gregor, D.~G. Reina, S.~L. Toral, An evolutionary approach to
  constrained path planning of an autonomous surface vehicle for maximizing the
  covered area of ypacarai lake, Soft Computing 23~(5) (2019) 1723--1734.

\bibitem{arzamendia2019comparison}
M.~Arzamendia, I.~Espartza, D.~G. Reina, S.~Toral, D.~Gregor, Comparison of
  eulerian and hamiltonian circuits for evolutionary-based path planning of an
  autonomous surface vehicle for monitoring ypacarai lake, Journal of Ambient
  Intelligence and Humanized Computing 10~(4) (2019) 1495--1507.

\bibitem{yanes2021dimensional}
S.~Yanes~Luis, D.~Guti{\'e}rrez-Reina, S.~Toral~Mar{\'\i}n, A dimensional
  comparison between evolutionary algorithm and deep reinforcement learning
  methodologies for autonomous surface vehicles with water quality sensors,
  Sensors 21~(8) (2021) 2862.

\bibitem{gul2021meta}
F.~Gul, W.~Rahiman, S.~Alhady, A.~Ali, I.~Mir, A.~Jalil, Meta-heuristic
  approach for solving multi-objective path planning for autonomous guided
  robot using pso--gwo optimization algorithm with evolutionary programming,
  Journal of Ambient Intelligence and Humanized Computing 12~(7) (2021)
  7873--7890.

\bibitem{carolina2022comparison}
M.~J. Ten~Kathen, I.~J. Flores, D.~G. Reina, A comparison of pso-based
  informative path planners for autonomous surface vehicles for water resource
  monitoring, in: 2022 7th International Conference on Machine Learning
  Technologies (ICMLT), 2022, pp. 271--276.

\bibitem{jara2022ola}
M.~J. Ten~Kathen, D.~G. Reina, I.~J. Flores, A comparison of pso-based
  informative path planners for detecting pollution peaks of the ypacarai lake
  with autonomous surface vehicles, in: International Conference on
  Optimization and Learning OLA’2022, \noop{3001}in press.

\bibitem{khan2021multimodal}
R.~A. Khan, S.~Yang, S.~Khan, S.~Fahad, et~al., A multimodal improved particle
  swarm optimization for high dimensional problems in electromagnetic devices,
  Energies 14~(24) (2021) 8575.

\bibitem{wang2013particle}
H.~Wang, W.~Wang, Z.~Wu, Particle swarm optimization with adaptive mutation for
  multimodal optimization, Applied Mathematics and Computation 221 (2013)
  296--305.

\bibitem{luo2020hybridizing}
W.~Luo, Y.~Qiao, X.~Lin, P.~Xu, M.~Preuss, Hybridizing niching, particle swarm
  optimization, and evolution strategy for multimodal optimization, IEEE
  Transactions on Cybernetics (2020).

\bibitem{zhang2020dynamic}
X.~T. Zhang, B.~Xu, W.~Zhang, J.~Zhang, X.~F. Ji, Dynamic neighborhood-based
  particle swarm optimization for multimodal problems, Mathematical Problems in
  Engineering 2020 (2020).

\bibitem{zhang2020modified}
X.~Zhang, H.~Liu, L.~Tu, A modified particle swarm optimization for multimodal
  multi-objective optimization, Engineering Applications of Artificial
  Intelligence 95 (2020) 103905.

\bibitem{chang2015modified}
W.~D. Chang, A modified particle swarm optimization with multiple
  subpopulations for multimodal function optimization problems, Applied Soft
  Computing 33 (2015) 170--182.

\bibitem{chang2017multimodal}
W.-D. Chang, Multimodal function optimizations with multiple maximums and
  multiple minimums using an improved pso algorithm, Applied Soft Computing 60
  (2017) 60--72.

\bibitem{zhang2019cluster}
W.~Zhang, G.~Li, W.~Zhang, J.~Liang, G.~G. Yen, A cluster based pso with leader
  updating mechanism and ring-topology for multimodal multi-objective
  optimization, Swarm and Evolutionary Computation 50 (2019) 100569.

\bibitem{zhang2013improved}
Y.~Zhang, X.~Xiong, Q.~Zhang, An improved self-adaptive pso algorithm with
  detection function for multimodal function optimization problems,
  Mathematical Problems in Engineering 2013 (2013).

\bibitem{khan2018modified}
S.~Khan, M.~Kamran, O.~U. Rehman, L.~Liu, S.~Yang, A modified pso algorithm
  with dynamic parameters for solving complex engineering design problem,
  International Journal of Computer Mathematics 95~(11) (2018) 2308--2329.

\bibitem{wang2012memetic}
H.~Wang, I.~Moon, S.~Yang, D.~Wang, A memetic particle swarm optimization
  algorithm for multimodal optimization problems, Information Sciences 197
  (2012) 38--52.

\bibitem{kennedy1995particle}
J.~Kennedy, R.~Eberhart, Particle swarm optimization, in: Proceedings of
  ICNN'95-international conference on neural networks, Vol.~4, IEEE, 1995, pp.
  1942--1948.

\bibitem{rasmussen2003gaussian}
C.~E. Rasmussen, Gaussian processes in machine learning, in: Summer school on
  machine learning, Springer, 2003, pp. 63--71.

\bibitem{mcmahan2016federated}
H.~B. McMahan, E.~Moore, D.~Ramage, B.~A. y~Arcas, Federated learning of deep
  networks using model averaging, arXiv preprint arXiv:1602.05629 2 (2016).

\bibitem{mcmahan2017communication}
B.~McMahan, E.~Moore, D.~Ramage, S.~Hampson, B.~A. y~Arcas,
  Communication-efficient learning of deep networks from decentralized data,
  in: Artificial intelligence and statistics, PMLR, 2017, pp. 1273--1282.

\bibitem{konevcny2016federated}
J.~Kone{\v{c}}n{\`y}, H.~B. McMahan, F.~X. Yu, P.~Richt{\'a}rik, A.~T. Suresh,
  D.~Bacon, Federated learning: Strategies for improving communication
  efficiency, arXiv preprint arXiv:1610.05492 (2016).

\bibitem{konevcny2016federatedb}
J.~Kone{\v{c}}n{\`y}, H.~B. McMahan, D.~Ramage, P.~Richt{\'a}rik, Federated
  optimization: Distributed machine learning for on-device intelligence, arXiv
  preprint arXiv:1610.02527 (2016).

\bibitem{sakai2018pythonrobotics}
A.~Sakai, D.~Ingram, J.~Dinius, K.~Chawla, A.~Raffin, A.~Paques,
  Pythonrobotics: a python code collection of robotics algorithms, arXiv
  preprint arXiv:1808.10703 (2018).

\bibitem{xin2019application}
J.~Xin, S.~Li, J.~Sheng, Y.~Zhang, Y.~Cui, Application of improved particle
  swarm optimization for navigation of unmanned surface vehicles, Sensors
  19~(14) (2019) 3096.

\bibitem{cui2020multi}
Y.~Cui, J.~Zhong, F.~Yang, S.~Li, P.~Li, Multi-subdomain grouping-based
  particle swarm optimization for the traveling salesman problem, IEEE Access 8
  (2020) 227497--227510.

\end{thebibliography}

\end{document}